\documentclass{article}

\usepackage[final]{nips_2016}

\usepackage[utf8]{inputenc} 
\usepackage[T1]{fontenc}    
\usepackage{float}
\usepackage{hyperref}       
\usepackage{url}            
\usepackage{booktabs}       
\usepackage{amsfonts}       
\usepackage{nicefrac}       
\usepackage{microtype}      

\usepackage{amsmath}
\usepackage{amsthm}
\usepackage{amssymb}
\usepackage{MnSymbol}
\usepackage{mathtools}
\usepackage{algorithm}
\usepackage{algpseudocode}
\usepackage{caption}
\makeatletter
\@addtoreset{algorithm}{section}
\makeatother

\newcommand{\R}[1]{\mathbb{R}^{#1}}
\newcommand{\lin}[2]{\mathcal{L}(#1; #2)}
\newcommand{\lefthook}{\righthalfcup}
\newcommand{\righthook}{\lefthalfcup}
\newcommand{\bignl}{S}
\newcommand{\smallnl}{\sigma}

\newcommand{\ip}[2]{\langle #1, \,#2 \rangle}
\newcommand{\tr}{\operatorname{tr}}

\newcommand{\dd}[2]{\frac{\mathrm{d} #1}{\mathrm{d} #2}}
\newcommand{\D}{\mathrm{D}}
\newcommand{\eval}[1]{\bigg|_{#1}}
\newcommand{\norm}[1]{\left\|#1\right\|}
\newcommand{\head}{\alpha}
\newcommand{\tail}{\omega}
\newcommand{\ind}{\xi}
\newcommand{\aelin}{\tau}

\newcommand\T{\rule{0pt}{2.6ex}}       
\newcommand\B{\rule[-1.2ex]{0pt}{0pt}} 

\newtheorem{definition}{Definition}[section]
\newtheorem{lemma}[definition]{Lemma}
\newtheorem{remark}[definition]{Remark}

\newtheorem{theorem}[definition]{Theorem}
\newtheorem{corollary}[definition]{Corollary}
\newtheorem{proposition}[definition]{Proposition}

\title{A Novel Representation of Neural Networks}

\author{
  Anthony L. Caterini and Dong Eui Chang \\
  Department of Applied Mathematics \\
  University of Waterloo \\
  Waterloo, ON, Canada, N2L 3G1 \\
  \texttt{\{alcateri, dechang\}@uwaterloo.ca}
}

\begin{document}

\maketitle

\begin{abstract}
Deep Neural Networks (DNNs) have become very popular for prediction in many areas. Their strength is in representation with a high number of parameters that are commonly learned via gradient descent or similar optimization methods. However, the representation is non-standardized, and the gradient calculation methods are often performed using component-based approaches that break parameters down into scalar units, instead of considering the parameters as whole entities. In this work, these problems are addressed. Standard notation is used to represent DNNs in a compact framework. Gradients of DNN loss functions are calculated directly over the inner product space on which the parameters are defined. This framework is general and is applied to two common network types: the Multilayer Perceptron and the Deep Autoencoder. 

\vspace{2mm}
{\bf Keywords:} Deep Learning, Neural Networks, Multilayer Perceptron, Deep Autoencoder, Backpropagation.
\end{abstract}


\section{Introduction}
Deep Neural Networks (DNNs) have grown increasingly popular over the last few years because of their astounding results in a variety of tasks. Their strength derives from their expressiveness, and this grows with network depth. However, the traditional approaches to representing DNNs suffers as the number of network layers increases. These often rely on confusing diagrams that provide an incomplete description of the mechanics of the network, which leads to complexity as the number of layers increases. Furthermore, DNNs are inconsistently formulated as a mathematical problem throughout research in the field, especially notationally, which impedes the efficiency in which results can be combined or expanded upon. A clear and concise framework underpinning DNNs must be developed, and this work endeavours to address that issue.

In this work, a novel mathematical framework for DNNs is created. It is formed by employing carefully selected standard notions and notation to represent a general DNN. Common mathematical tools such as the inner product, the adjoint operation, and maps defined over generic inner product spaces are utilized throughout this work. Well-established mathematical objects are treated as-is in this framework; it is no longer necessary to convert a matrix into a column vector or a decompose it into a collection of components, for example, for the purposes of derivative calculation. This work presents a comprehensive mathematical standard upon which DNNs can be formulated. 

The specific layout of this paper is as follows. After some mathematical preliminaries, a generic DNN is formulated over an abstract inner product space. The chain rule is used to demonstrate a concise coordinate-free approach to backpropagation. Two standard loss functions are explicitly considered, and it is shown how to handle some variations on those within the learning algorithm. Then, this framework is applied to the multilayer perceptron (MLP). The specifics of the previous approach become clear, and it is shown how to create a gradient descent algorithm to learn the parameters of the MLP. Some of the theory developed in the section on MLP is then applied to a deep autoencoder (AE), which demonstrates the flexibility of the approach. This type of framework can be extended to other types of networks, including convolutional neural networks (CNNs) and recurrent neural networks (RNNs), but these are omitted for the sake of brevity. 

\section{Mathematical Preliminaries}
In this section, we set notation and review some elementary but essential mathematical facts. These facts will be used to cast neural networks into a novel framework in the following sections.

\subsection{Linear Maps, Bilinear Maps, and Adjoints} \label{sec:bilinear}
Consider three  inner product spaces $E_1, E_2,$ and $E_3$, i.e. each vector space is equipped with an inner product denoted by $\ip{\,}{}$. The space of linear maps from $E_1$ to $E_2$ will be denoted $\lin{E_1}{E_2}$. Note that for $L \in \lin{E_1}{E_2}$ and $u \in E_1$, $L \cdot u \in E_2$ denotes $L$ operating on $u$, i.e. $L(u)$ or more simply $Lu$. Similarly, the space of bilinear maps from $E_1 \times E_2$ into $E_3$ will be denoted $\lin{E_1, E_2}{E_3}$. For $B \in \lin{E_1, E_2}{E_3}$ and $u_1 \in E_1, u_2 \in E_2$, $B \cdot (u_1, u_2) \in E_3$ denotes $B$ operating on $u_1$ and $u_2$, i.e. $B(u_1, u_2)$. For any bilinear map $B \in \lin{E_1, E_2}{E_3}$ and any $e_1 \in E_1$, a linear map $e_1 \lefthook B \in  \lin{E_2}{E_3}$ is defined as follows: 
\[
(e_1  \lefthook B) \cdot  e_2 = B(e_1,e_2)
\]
for all $e_2 \in E_2$.
Similarly, for any $e_2 \in E_2$, a linear map $ B \righthook e_2  \in  \lin{E_1}{E_3}$ is defined as follows: 
\[
( B \righthook e_2)  \cdot e_1 = B(e_1,e_2).
\]
for all $e_1 \in E_1$. These operators $ \lefthook $ and $\righthook$ will be referred to as the \emph{left hook} and \emph{right hook} operators, respectively.

The adjoint $L^*$ of a linear map $L \in \lin{E_1}{E_2}$ is a linear map in $\lin{E_2}{E_1}$ defined by
\[
\langle L^* e_2, e_1 \rangle = \langle e_2, Le_1\rangle
\]
for all $e_1 \in E_1$ and $e_2\in E_2$.  The adjoint operator satisfies the direction reversing property:
\[
(L_2 L_1)^* = L_1^* L_2^*
\]
for all $L_1 \in \lin{E_1}{E_2}$ and $L_2 \in \lin{E_2}{E_3}$.
\subsection{Derivatives}
In this section, notation for derivatives in accordance with \cite{marsden1988manifolds} is presented. 

\subsubsection{First Derivatives} \label{sec:first_deriv}
Consider a map $f : E_1 \rightarrow E_2$, where $E_1$ and $E_2$ are inner product spaces. The (first) derivative map of $f$, denoted $\D f$, is a map from $E_1$ to $\lin{E_1}{E_2}$ that operates as $x \mapsto \D f(x)$ for any $x \in E_1$. The linear map $\D f(x)$ operates in the following manner for any $v \in E_1$:
\begin{align} \label{eqn:Df}
\D f(x) \cdot v = \left. \dd{}{t} f(x + t v) \right|_{t=0}.
\end{align}
For each $x \in E_1$ the adjoint of the derivative $\D f(x) \in  \lin{E_1}{E_2}$ is well defined with respect to the inner products on $E_1$ and $E_2$, and it  is denoted  $\D^* f(x)$ instead of $\D  f(x)^*$ for the sake of notational convenience. Then, $\D^*f : E_1 \rightarrow \lin{E_2}{E_1}$ denotes the adjoint map that maps each point $x \in E_1$ to $\D^* f(x)\in \lin{E_2}{E_1}$.

Now consider two maps $f_1 : E_1 \rightarrow E_2$ and $f_2 : E_2 \rightarrow E_3$, where $E_3$ is another inner product space. The derivative of their composition, $\D (f_2 \circ f_1)(x) \in \lin{E_1}{E_3}$ for $x \in E_1$, is calculated using the well-known chain rule.
\begin{lemma}[Chain Rule]
For any $x \in E_1$,
\[
\D (f_2 \circ f_1) (x) = \D f_2(f_1(x)) \cdot \D f_1(x),
\]
where $f_1 : E_1 \rightarrow E_2$ and $f_2 : E_2 \rightarrow E_3$ are $C^1$, i.e. continuously differentiable, and $E_1, E_2, $ and $E_3$ are vector spaces.
\end{lemma}

\subsubsection{Second Derivatives}
Every map in here assumed to be (piecewise) $C^2$, i.e. (piecewise) twice continuously differentiable, unless stated otherwise. 
The second derivative map of $f$, denoted $\D^2 f$, is a map from $E_1$ to $\lin{E_1, E_1}{E_2}$, which operates as $x \mapsto \D^2f(x)$ for any $x \in E_1$. The bilinear map $\D^2f(x)$ operates as follows: for any $v_1, v_2 \in E_1$
\begin{align} \label{eqn:D2f}
\D^2f(x) \cdot (v_1, v_2) = \left. \dd{}{t} \left(\D f(x + tv_1) \cdot v_2\right) \right|_{t=0}.
\end{align}
It is not hard to show that $\D^2f(x)$ is symmetric, i.e. $\D^2f(x) \cdot (v_1, v_2) = \D^2f(x) \cdot (v_2, v_1)$ for all $v_1, v_2 \in E_1$. Furthermore, it can be shown that
\begin{align*}
\D^2f(x) \cdot (v_1, v_2) = \left. \frac{\partial^2}{\partial t \partial s} f(x + tv_1 + sv_2) \right|_{t = s = 0}.
\end{align*}
The hook notation from Section \ref{sec:bilinear} can be used to turn the second derivative into a linear map. In particular, $(v \lefthook \D^2f(x))$ and $(\D^2f(x) \righthook v) \in \lin{E_1}{E_2}$ for any $x, v \in E_1$. An important identity exists for the second derivative of the composition of two functions.

\begin{lemma} \label{lem:chain_rule_2}
For any $x, v_1, v_2 \in E_1$, 
\[
\D^2 (f_2 \circ f_1) (x) \cdot (v_1, v_2) = \D^2 f_2(f_1(x)) \cdot (\D f_1(x) \cdot v_1, \D f_1(x) \cdot v_2) + \D f_2(f_1(x)) \cdot \D^2 f_1(x) \cdot (v_1, v_2),
\]
where $f_1 : E_1 \rightarrow E_2$ is $C^1$ and  $f_2 : E_2 \rightarrow E_3$ is  $C^2$ for vector spaces $E_1, E_2,$ and $E_3$. 
\end{lemma}
This can be seen as the chain rule for second derivatives. 

\subsubsection{Parameter-Dependent Maps}
Now suppose $f$ is a map from $E_1 \times H_1 \rightarrow E_2$, i.e. $f(x; \theta) \in E_2$ for any $x \in E_1$ and $\theta \in H_1$, where $H_1$ is also an inner product space. The variable $x \in E_1$ is said to be the \emph{state variable} for $f$, whereas $\theta \in H_1$ is a \emph{parameter}. The notation presented in \eqref{eqn:Df} is used to denote the derivative of $f$ with respect to the state variable, i.e. for all $v \in E_1$,
\[
\D f(x; \theta) \cdot v = \left. \dd{}{t} f(x + t v; \theta) \right|_{t=0}.
\]
Also, $\D^2f(x;\theta) \cdot (v_1, v_2) = \D\left(\D f(x;\theta) \cdot v_2\right) \cdot v_1$ as before. 
New notation is used to denote the derivative of $f$ with respect to the parameters, as follows:
\begin{align*}
\nabla f(x;\theta) \cdot u = \left. \dd{}{t} f(x; \theta + tu) \right|_{t=0}
\end{align*}
for any $u \in H_1$. Note that $\nabla f(x;\theta) \in \lin{H_1}{E_2}$. In the case where $f$ depends on two parameters as $f(x; \theta_1, \theta_2)$, the notation $\nabla_{\theta_1} f(x; \theta_1, \theta_2)$  will be used to explicitly denote differentiation with respect to the parameter $\theta_1$  when the distinction is necessary.

The mixed partial derivative maps, $\nabla \D f(x; \theta) \in \lin{H_1, E_1}{E_2}$ and $\D\nabla f(x;\theta) \in \lin{E_1, H_1}{E_2}$, are defined as:  
\begin{align*}
\nabla \D f(x;\theta) \cdot (u, e) &= \left. \dd{}{t} \left(\D f(x; \theta + tu) \cdot e\right) \right|_{t=0}, \\
\D \nabla f(x; \theta) \cdot (e, u) &= \left. \dd{}{t} \left(\nabla f(x + te; \theta) \cdot u\right) \right|_{t=0}.
\end{align*}
for any $e \in E_1$, $u \in H_1$. Note that if $f \in C^2$, then $\D \nabla f(x;\theta) \cdot (u, e) = \nabla \D f(x;\theta) \cdot (e, u)$, i.e. the mixed partial derivatives are equal. 

\subsection{Elementwise Functions}
Consider an inner product space $E$ of dimension $n$ with the inner product denoted by $\ip{\,}{}$. Let $\{e_k\}_{k=1}^n$ be an orthonormal basis of $E$. An \emph{elementwise function} is defined to be a function $\Psi : E \rightarrow E$ of the form
\begin{align} \label{eqn:elem_fn}
\Psi(v) = \sum_{k=1}^n \psi(\ip{v}{e_k})e_k,
\end{align}
where $\psi : \R{} \rightarrow \R{}$ --- known as the \emph{elementwise operation} associated with $\Psi$ --- defines the operation of the elementwise function over the components $\{\ip{v}{e_k}\}_k$ of the vector $v \in E$. The operator $\Psi$ is basis-dependent, but $\{e_k\}_{k=1}^n$ can be any orthonormal basis of $E$. Also define the \emph{elementwise first derivative} of an elementwise function $\Psi$, $\Psi' : E \rightarrow E$, as 
\begin{align} \label{eqn:elem_first_deriv}
\Psi'(v) = \sum_{k=1}^n \psi'(\ip{v}{e_k})e_k,
\end{align}
where $\psi'$ is the first derivative of $\psi$. Note that $\psi'$ can be referred to as the associated elementwise operation for $\Psi'$. Similarly, define the \emph{elementwise second derivative} function $\Psi'': E \rightarrow E$ as 
\begin{align} \label{eqn:elem_second_deriv}
\Psi''(v) = \sum_{k=1}^n \psi''(\ip{v}{e_k}) e_k,
\end{align}
where $\psi''$ is the second derivative of $\psi$. 

\subsubsection{Hadamard Product}
Now define a symmetric bilinear operator $\odot \in \lin{E, E}{E}$ over the basis vectors $\{e_k\}_{k=1}^n$ as
\begin{align} \label{eqn:hadamard}
e_k \odot e_{k'} \coloneqq \delta_{k,k'} e_k,
\end{align}
where $\delta_{k,k'}$ is the Kronecker delta. This is the standard Hadamard product when $E = \R{n}$ and $\{e_k\}_{k=1}^n$ is the standard basis of $\R{n}$. However, when $E \neq \R{n}$ or $\{e_k\}_{k=1}^n$ is not the standard basis, $\odot$ can be seen as a generalization of the Hadamard product, and it will be referred to as such in this paper. For illustrative purposes, consider the (generalized) Hadamard product of two vectors $v, v' \in E$. These vectors can be written as $v = \sum_{k=1}^n \ip{v}{e_k} e_k$ and $v' = \sum_{k=1}^n \ip{v'}{e_k} e_k$.  Then,
\begin{align*}
v \odot v' &= \left(\sum_{k=1}^n \ip{v}{e_k}e_k\right) \odot \left(\sum_{k'=1}^n \ip{v'}{e_{k'}}e_{k'}\right) \\
&= \sum_{k,k'=1}^n \ip{v}{e_k} \ip{v'}{e_{k'}} \left(e_k \odot e_{k'}\right) \\
&= \sum_{k=1}^n \ip{v}{e_k} \ip{v'}{e_k} e_k.
\end{align*}
It is easy to show that the Hadamard product  satisfies the following properties:
\begin{align*}
&v \odot v' = v' \odot v,\\
&(v \odot v' ) \odot y = v \odot ( v'  \odot y ),\\ 
&\langle y, v \odot v' \rangle = \langle v \odot y, v'\rangle = \langle y \odot v', v\rangle
\end{align*}
for all $y, v, v' \in E$.

\subsubsection{Derivatives of Elementwise Functions}
Some results regarding the derivative maps for a generic elementwise function $\Psi$, i.e. $\D\Psi$ and $\D^2\Psi$, are presented now. 

\begin{proposition}
\label{prop:DS}
Let $\Psi : E \rightarrow E$ be an elementwise function as defined in \eqref{eqn:elem_fn}, for an inner product space $E$ of dimension $n$ with a basis $\{e_k\}_{k=1}^n$ and inner product $\ip{\,}{}$. Then, for any $v, z \in E$,
\[
\D\Psi(z) \cdot v = \Psi'(z) \odot v, 
\]
where the Hadamard product $\odot$ is defined in \eqref{eqn:hadamard} and $\Psi'$ is the elementwise first derivative defined in \eqref{eqn:elem_first_deriv}. Furthermore, $\D\Psi(z)$ is self-adjoint, i.e. $\D^*\Psi(z) = \D\Psi(z)$ for all $z\in E$.
\begin{proof} Let $\psi$ be the elementwise operation associated with $\Psi$. Then,
\begin{align*}
\D\Psi(z) \cdot v &= \dd{}{t}\Psi(z + tv)\eval{t=0} \\
&= \dd{}{t} \sum_{k=1}^n \psi(\ip{z + tv}{e_k}) e_k \eval{t=0} \\
&= \sum_{k=1}^n \psi'(\ip{z}{e_k}) \ip{v}{e_k} e_k \\
&= \Psi'(z) \odot v,
\end{align*}
where the third equality follows from the chain rule and linearity of the derivative. 

Furthermore, let $y \in E$. Then,
\begin{align*}
\ip{y}{\D\Psi(z)\cdot v} &= \ip{y}{\Psi'(z) \odot v} \\
&= \ip{\Psi'(z) \odot y}{v} \\
&= \ip{\D\Psi(z) \cdot y}{v}.
\end{align*}
Since $\ip{y}{\D\Psi(z)\cdot v} = \ip{\D\Psi(z) \cdot y}{v}$ for any $v, y, z \in E$, $\D\Psi(z)$ is self-adjoint.
\end{proof}
\end{proposition}

\begin{proposition} \label{prop:DtwoS}
Let $\Psi : E \rightarrow E$ be an elementwise function as defined in \eqref{eqn:elem_fn}, for an inner product space $E$ of dimension $n$ with a basis $\{e_k\}_{k=1}^n$ and inner product $\ip{\,}{}$. Then, for any $v_1, v_2, z \in E$, 
\begin{align} \label{eqn:D2S}
\D^2\Psi(z) \cdot (v_1, v_2) = \Psi''(z) \odot v_1 \odot v_2,
\end{align}
where the Hadamard product $\odot$ is defined in \eqref{eqn:hadamard} and $\Psi''$ is the elementwise second derivative defined in \eqref{eqn:elem_second_deriv}. Furthermore, $\left(v_1 \lefthook \D^2\Psi(z)\right)$ and $\left(\D^2\Psi(z) \righthook v_2 \right)$ are both self-adjoint linear maps  for any $v_1, v_2, z \in E$. 
\begin{proof} Prove \eqref{eqn:D2S} directly: 
\begin{align*}
\D^2\Psi(z) \cdot (v_1, v_2) &= \D (\D\Psi(z) \cdot v_2) \cdot v_1 \\
&= \D (\Psi'(z) \odot v_2) \cdot v_1 \\
&= (\Psi''(z) \odot v_1) \odot v_2,
\end{align*}
where the third equality follows since $\Psi'(z)\odot v_2$ is an elementwise function in $z$. Also, for any $y \in E$, 
\begin{align*}
\ip{y}{\left(v_1 \lefthook \D^2\Psi(z)\right) \cdot v_2} &= \ip{y}{\D^2\Psi(z) \cdot (v_1, v_2)} \\
&= \ip{y}{\Psi''(z) \odot v_1 \odot v_2} \\
&= \ip{\Psi''(z) \odot v_1 \odot y}{v_2}\\
&= \ip{\left(v_1 \lefthook \D^2\Psi(z) \right) \cdot y}{v_2}.
\end{align*}
This implies that $\left(v_1 \lefthook \D^2\Psi(z)\right)$ is self-adjoint for any $v_1, z \in E$. Since $\D^2\Psi(z)$ is a symmetric bilinear map, this also implies that $\left(\D^2\Psi(z) \righthook v_1\right)$ is self-adjoint for any $v_1, z \in E$.
\end{proof}
\end{proposition} 

\section{Coordinate-Free Representation of  Neural Networks} \label{sec:main_formulation}
In this section, coordinate-free backpropagation is derived for a generic layered neural network. The network is formulated and then a gradient descent algorithm is given for two types of loss functions.  

\subsection{Neural Network Formulation}
Neural networks are layered models, with the actions of layer $i$ denoted by $f_i: E_i \times H_i \rightarrow E_{i+1}$, where $E_i, H_i,$ and $E_{i+1}$ are inner product spaces. In other words, $f_i(x_i, \theta_i) \in E_{i+1}$ for $x_i \in E_i$ and  $\theta_i \in H_i$. For a neural network with $L$ layers, $i \in \{1, \ldots, L\}$. The \emph{state variable} $x_i \in E_i$ is an abstract representation of the input data $x_1 = x$ at layer $i$. The \emph{parameters} $\theta_i \in H_i$ at layer $i$ must be learned, often by some form of gradient descent. Note that the explicit dependence of $f_i$ on the parameter $\theta_i$ will be suppressed in the notation throughout this section. In this way, $f_i : E_i \rightarrow E_{i+1}$, defined by $x_{i+1} = f_i(x_i)$, where $f_i$ \emph{depends} on $\theta_i$. Then, the network prediction can be written as a composition of functions 
\begin{align} \label{eqn:F}
F(x; \theta) =( f_L \circ \cdots \circ f_1)(x),
\end{align} 
where each $f_i : E_i \rightarrow E_{i+1}$ has a suppressed dependence on the parameter $\theta_i \in H_i$, and $\theta$ represents the parameter set $\{\theta_1, \ldots, \theta_L\}$. Each parameter $\theta_i$ is independent of the other parameters $\{\theta_j\}_{j\neq i}$ in this formulation. 

Some maps will be introduced to assist in derivative calculation. Let the \emph{head} map at level $i$, $\head_i : E_1 \rightarrow E_{i+1}$, be defined by:
\begin{align} \label{eqn:head}
\head_i = f_i \circ \cdots \circ f_1
\end{align}
for each  $i \in \{1, \ldots, L\}$. Note that $\head_i$ implicitly depends on the parameters $\{\theta_1, \ldots, \theta_i\}$. For convenience, set $\head_0$ to be the identity map on $E_1$. Similarly, define the \emph{tail} map at level $i$, $\tail_i : E_i \rightarrow E_{L+1}$, as:
\begin{align} \label{eqn:tail}
\tail_i = f_L \circ \cdots \circ f_i
\end{align}
for each $i \in \{1, \ldots, L\}$. The map $\tail_i$ implicitly depends on $\{\theta_i, \ldots, \theta_L\}$. Again for convenience, set $\tail_{L+1}$ to be the identity map on $E_{L+1}$. It is easy to show that the following hold for all $i \in \{1, \ldots, L\}$:
\begin{equation}\label{eqn:recursive_rel}
F = \tail_{i+1} \circ \head_i, \quad \tail_i = \tail_{i+1} \circ f_i, \quad \head_i = f_i \circ \head_{i-1}.
\end{equation}
The equations in \eqref{eqn:recursive_rel} imply that the prediction $F$ can be decomposed into 
\[
F = \tail_{i+1} \circ f_i \circ \head_{i-1}
\]
for all $i \in \{1, \ldots, L\}$, where $\head_{i-1}$ does not depend on the parameter $\theta_i$.

\subsection{Loss Function and Backpropagation} \label{sec:standard_loss}
While training a neural network, the goal is to optimize some loss function $J$ with respect to the parameters $\theta$. For example, consider 
\begin{align}
J(x; \theta) \coloneqq \frac{1}{2} \norm{y - F(x; \theta)}^2 = \frac{1}{2} \ip{y - F(x; \theta)}{y - F(x; \theta)}, \label{eqn:standard_loss}
\end{align}
where $y \in E_{L+1}$ is the known response data. Gradient descent is used to optimize the loss function, thus the gradient of $J$ with respect to each of the parameters must be calculated. Before that can be done, some preliminary results will be introduced. In this section, it is always assumed that 
\[
x_i = \head_{i-1}(x)
\]
 is the state variable at level $i$ for a given data point $x$.
\begin{theorem}\label{theorem:nabla:F:general} Let $J$ be defined as in \eqref{eqn:standard_loss}. Then, for any $x \in E_1$ and $i \in \{1, \ldots, L\}$, 
\begin{align} \label{eqn:dJdT}
\nabla_{\theta_i}J(x; \theta) = \nabla^*_{\theta_i}F(x; \theta) \cdot (F(x; \theta) - y).
\end{align}
\begin{proof}
By the product rule, for any $U_i \in H_i$,
\begin{align*}
\nabla_{\theta_i}J(x; \theta) \cdot U_i &= \ip{F(x; \theta) - y}{\nabla_{\theta_i} F(x; \theta) \cdot U_i} = \ip{\nabla^*_{\theta_i}F(x; \theta) \cdot (F(x; \theta) - y)}{U_i}. 
\end{align*}
Since this holds for any $U_i \in H_i$, \eqref{eqn:dJdT} follows.
\end{proof}
\end{theorem}
The following two theorems show how to  compute the derivative $\nabla_{\theta_i}J(x; \theta)$ given in \eqref{eqn:dJdT} recursively.
\begin{theorem}\label{theorem:nabla_F}
With $F$ defined as in \eqref{eqn:F} and $\tail_i$ defined as in \eqref{eqn:tail},
\begin{align}\label{nabla:star:F}
\nabla^*_{\theta_i} F(x; \theta) = \nabla^*_{\theta_i}f_i(x_i) \cdot \D^*\tail_{i+1}(x_{i+1})
\end{align}
with $x_i = \head_{i-1}(x)$ and $x_{i+1} = f_i(x_i)$, for all $i \in \{1, \ldots, L\}$.
\begin{proof}
Apply the chain rule to  $F = \tail_{i+1} \circ f_i \circ \head_{i-1}$ and then take the adjoint of it to get the result.
\end{proof}
\end{theorem}

\begin{theorem} \label{theorem:backprop} With $\tail_i$ defined as in \eqref{eqn:tail}, then for all $x_i \in E_i$, 
\begin{align}\label{Dtail:rec}
\D\tail_i(x_i) = \D\tail_{i+1}(x_{i+1}) \cdot \D f_i(x_i)
\end{align}
and
\begin{align}\label{backprop:general:omega}
\D^*\tail_i(x_i) = \D^* f_i(x_i) \cdot \D^*\tail_{i+1}(x_{i+1}),
\end{align}
where $x_{i+1} = f_i(x_i)$, for all $i \in \{1, \ldots, L\}$.
\begin{proof}
Apply the chain rule to $\tail_i (x_i)= (\tail_{i+1} \circ f_i)(x_i)$ to get \eqref{Dtail:rec}. 
Then, take the adjoint of \eqref{Dtail:rec} to get \eqref{backprop:general:omega}. This holds for any $i \in \{1, \ldots, L\}$. 
\end{proof}
\end{theorem}

Algorithm \ref{alg:general_grad_desc} provides a method to perform one iteration of gradient descent to minimize $J$ over the parameter set $\theta = \{\theta_1, \ldots, \theta_L\}$ for a single data point $x$. The algorithm extends linearly to a batch of updates over multiple data points. Notice that gradient descent is performed directly over the inner product space $H_i$ at each layer $i$, which contrasts the standard approach of performing the descent over each individual component of $\theta_i$. This can be seen as a coordinate-free gradient descent algorithm.

\begin{algorithm}
\caption{One iteration of gradient descent for a general NN}
\label{alg:general_grad_desc}
\begin{algorithmic}
\Function{Descent Iteration}{$x, y,\theta_1, \ldots, \theta_L, \eta$}
\State $x_1 \gets x$
\For {$i \in \{1, \ldots, L\}$} \Comment $x_{L+1} = F(x; \theta)$
\State $x_{i+1} \gets f_i(x_i)$
\EndFor
\For {$i \in \{L, \ldots, 1\}$}
\State $\tilde{\theta_i} \gets \theta_i$ \Comment Store old $\theta_i$ for updating $\theta_{i-1}$
\If {$i = L$} \Comment $e = \D^*\tail_{i+1}(x_{i+1}) \cdot \left(x_{L+1} - y\right)$
\State $e \gets x_{L+1} - y$   \Comment $\tail_{L+1} = \textup{identity}$ 
\Else
\State $e \gets \D^*f_{i+1}(x_{i+1}) \cdot e$ \Comment \eqref{backprop:general:omega}, update with $\tilde{\theta}_{i+1}$  
\EndIf
\State $\nabla_{\theta_i} J(x; \theta) \gets \nabla_{\theta_i}^* f_i(x_i) \cdot e$ \Comment Thms. \ref{theorem:nabla:F:general} and \ref{theorem:nabla_F}
\State $\theta_i \gets \theta_i - \eta \nabla_{\theta_i} J(x; \theta)$
\EndFor
\EndFunction
\end{algorithmic}
\end{algorithm}

\begin{remark} \label{rem:l2_reg}
It is not difficult to incorporate a standard $\ell_2$-regularizing term into this framework. Construct a new objective function $\mathcal{J}_T(x; \theta) = J(x; \theta) + \lambda T(\theta)$, where $\lambda \in \R{}_{\geq 0}$ is the \emph{regularization parameter} and 
\[ T(\theta) = \frac{1}{2} \norm{\theta}^2 = \frac{1}{2} \sum_{i=1}^L \norm{\theta_i}^2 = \frac{1}{2} \sum_{i=1}^L \ip{\theta_i}{\theta_i} \]
is the \emph{regularization term}. It follows that $\nabla_{\theta_i} \mathcal{J}_T(x; \theta) = \nabla_{\theta_i} J(x; \theta) + \lambda \theta_i$, since $\nabla_{\theta_i} T(\theta) = \theta_i$. This implies that gradient descent can be updated to include the regularizing term, i.e. the last line in Algorithm \ref{alg:general_grad_desc} can be altered as follows:
\[
\theta_i \leftarrow \theta_i - \eta \left(\nabla_{\theta_i} J(x; \theta) + \lambda \theta_i\right).
\]
\end{remark}

\begin{remark} The loss function considered so far was $J(x;\theta) = \frac{1}{2} \norm{y - F(x;\theta)}^2$. However, another standard loss function is the cross-entropy loss, 
\[
\tilde{J}(x; \theta) = - \ip{y}{L(F(x;\theta))} - \ip{\mathbf{1} - y}{L(\mathbf{1} - F(x;\theta))},
\]
where $\mathbf{1}$ is a vector of ones of appropriate length and $L$ is an elementwise function with elementwise operation $\log$. The gradient of $\tilde{J}$ with respect to a parameter $\theta_i$, in the direction of $U_i$, is 
\begin{align*}
\nabla_{\theta_i} \tilde{J}(x;\theta)\! \cdot\! U_i\! &= - \ip{y}{\D L(F(x;\theta))\! \cdot \!\nabla_{\theta_i} F(x;\theta)\! \cdot\! U_i} + \ip{\mathbf{1} \!-\! y}{\D L(\mathbf{1} \!-\! F(x;\theta))\! \cdot\! \nabla_{\theta_i} F(x;\theta) \cdot U_i} \\
&= \ip{\nabla_{\theta_i}^*F(x;\theta) \cdot \left[- \D L(F(x;\theta)) \cdot y + \D L(\mathbf{1} - F(x;\theta)) \cdot (\mathbf{1} - y)\right]}{U_i}.
\end{align*}
Thus, 
\[
\nabla_{\theta_i} \tilde{J}(x;\theta) = \nabla_{\theta_i}^*F(x;\theta) \cdot \left[- \D L(F(x;\theta)) \cdot y + \D L(\mathbf{1} - F(x;\theta)) \cdot (\mathbf{1} - y)\right].
\]
 Algorithm \ref{alg:general_grad_desc} can then be modified to minimize $\tilde{J}$ instead of $J$ by changing the initialization of the error $e$ from $e \gets x_{L+1} - y$ to $e \gets -\D L(F(x;\theta)) \cdot y + \D L(\mathbf{1} - F(x;\theta)) \cdot (\mathbf{1} - y)$. 
\end{remark}

\subsection{Higher-Order Loss Function}
Suppose that another term is added to the loss function to penalize the first order derivative of $F(x;\theta)$, as in \cite{rifai2011manifold} or \cite{simard1992tangent} for example. This can be represented using 
\begin{align} \label{eqn:R}
R(x; \theta) \coloneqq \frac{1}{2} \norm{\D F(x; \theta) \cdot v_x - \beta_x}^2,
\end{align}
for some $v_x \in E_1$ and $\beta_x \in E_{L+1}$. When $\beta_x = 0$, minimizing $R(x; \theta)$ promotes invariance of the network in the direction of $v_x$. Similarly to Remark \ref{rem:l2_reg}, $R$ can be added to $J$ to create a new loss function 
\begin{align} \label{eqn:curly_J}
\mathcal{J}_R(x;\theta) = J(x; \theta) + \mu R(x; \theta),
\end{align}
where $\mu \in \R{}_{\geq 0}$ determines the amount that the higher-order term contributes to the loss function. Note that $R$ can be extended additively to contain multiple terms:
\begin{align} \label{eqn:R_mult_terms}
R(x; \theta) = \sum_{(v_x, \beta_x) \in \mathcal B_x} \frac{1}{2} \norm{\D F(x; \theta) \cdot v_x - \beta_x}^2,
\end{align}
where $\mathcal B_x$ is a finite set of pairs $(v_x, \beta_x)$ for each data point $x$.

\begin{theorem} \label{theorem:nabla:theta;R} For all $i \in \{1, \ldots, L\}$, with $R$ defined as in \eqref{eqn:R},
\begin{align}  \label{eqn:dRdW}
\nabla_{\theta_i}R(x; \theta) = \left(\nabla_{\theta_i} \D F(x; \theta) \righthook v_x\right)^* \cdot \left(\D F(x; \theta) \cdot v_x - \beta_x\right).
\end{align}  
\begin{proof}
From \eqref{eqn:R}, 
\begin{align*}
\nabla_{\theta_i}R(x; \theta) \cdot U_i &= \ip{\D F(x; \theta) \cdot v_x - \beta_x}{\nabla_{\theta_i} \D F(x; \theta) \cdot (U_i, v_x)}\\
&= \ip{\D F(x; \theta) \cdot v_x - \beta_x}{\left(\nabla_{\theta_i} \D F(x; \theta) \righthook v_x\right) \cdot U_i} \\
&=\langle  \left(\nabla_{\theta_i} \D F(x; \theta) \righthook v_x\right)^* \cdot \left(\D F(x; \theta) \cdot v_x - \beta_x\right), U_i \rangle
\end{align*}
for all $U_i \in H_i, i \in \{1, \ldots, L\}$. Thus, \eqref{eqn:dRdW} follows.
\end{proof}
\end{theorem}

Some preliminary results will be given before \eqref{eqn:dRdW} can be recursively computed. Note again in this section that 
\[
x_i = \head_{i-1}(x)
\]
 is the state variable at layer $i$ for input data $x$, where the map $\head_i$ is defined in \eqref{eqn:head}. 

\begin{lemma} \label{lem:forward_prop} For any $x \in E_1$, and $i \in \{1, \ldots, L\}$, 
\[
\D\head_i(x) = \D f_i(x_i) \cdot \D\head_{i-1}(x).
\]

\begin{proof}
This is proven using the chain rule, since $\head_i = f_i \circ \head_{i-1}$ for all $i \in \{1, \ldots, L\}$.
\end{proof}
\end{lemma}
Lemma \ref{lem:forward_prop} defines forward propagation through the tangent network, in the spirit of \cite{simard1992tangent}. Note that since $\head_L = F$, $\D\head_L = \D F$. This implies that Lemma \ref{lem:forward_prop} is needed for calculating $\D F(x; \theta) \cdot v_x$. Now, tangent backpropagation will be described.

\begin{theorem}[Tangent Backpropagation] \label{thm:tgt_backprop}
For any $x, v \in E_1$, and $i \in \{1, \ldots, L\}$, 
\begin{align*}
\left(\left(\D \head_{i-1}(x) \cdot v \right) \lefthook \D^2\tail_i(x_i)\right)^* &= \D^*f_i(x_i) \cdot \left((\D \head_i(x) \cdot v) \lefthook \D^2\tail_{i+1}(x_{i+1}) \right)^* \\
&\qquad + \left(\left(\D \head_{i-1}(x) \cdot v \right) \lefthook \D^2 f_i(x_i)\right)^* \cdot \D^*\tail_{i+1}(x_{i+1}),
\end{align*}
where $\head_i$ is defined in \eqref{eqn:head} and $\tail_i$ is defined in \eqref{eqn:tail}.

\begin{proof} Let $v_1, v_2,$ and $z \in E_i$. Then, 
\begin{align*}
\left(v_1 \lefthook \D^2\tail_i(z)\right) \cdot v_2 &= \D^2\tail_i(z) \cdot (v_1, v_2) \\
&= \D^2(\tail_{i+1} \circ f_i)(z) \cdot (v_1, v_2) \\
&= \D^2 \tail_{i+1}(f_i(z)) \cdot \left(\D f_i(z) \cdot v_1, \D f_i(z) \cdot v_2\right) \\
&\qquad + \D\tail_{i+1}(f_i(z)) \cdot \D^2 f_i(z) \cdot (v_1, v_2) \\
&= \left((\D f_i(z) \cdot v_1) \lefthook \D^2\tail_{i+1}(f_i(z)) \right)\cdot \D f_i(z) \cdot v_2 \\
&\qquad + \D\tail_{i+1}(f_i(z)) \cdot \left(v_1 \lefthook \D^2 f_i(z)\right) \cdot v_2,
\end{align*}
where the third equality comes from Lemma \ref{lem:chain_rule_2}. The operator $\left(v_1 \lefthook \D^2\tail_i(z)\right)$ can thus be written as
\[
\left(v_1 \lefthook \D^2\tail_i(z)\right) = \left((\D f_i(z) \cdot v_1) \lefthook \D^2\tail_{i+1}(f_i(z)) \right)\cdot \D f_i(z) + \D\tail_{i+1}(f_i(z)) \cdot \left(v_1 \lefthook \D^2 f_i(z)\right).
\]
By taking the adjoint,
\[
\left(v_1 \lefthook \D^2\tail_i(z)\right)^* = \D^*f_i(z) \cdot \left((\D f_i(z) \cdot v_1) \lefthook \D^2\tail_{i+1}(f_i(z)) \right)^*  + \left(v_1 \lefthook \D^2 f_i(z)\right)^* \cdot \D^*\tail_{i+1}(f_i(z)).
\]
Set $v_1 = \D \head_{i-1}(x) \cdot v$ and $z = x_i$ to obtain the final result:
\begin{align*}
\left(\left(\D \head_{i-1}(x) \cdot v \right) \lefthook \D^2\tail_i(x_i)\right)^* &= \D^*f_i(x_i) \cdot \left((\D \head_i(x) \cdot v) \lefthook \D^2\tail_{i+1}(x_{i+1}) \right)^* \\
&\qquad + \left(\left(\D \head_{i-1}(x) \cdot v \right) \lefthook \D^2 f_i(x_i)\right)^* \cdot \D^*\tail_{i+1}(x_{i+1}),
\end{align*}
where $\D \head_i(x) \cdot v = \D f_i(x_i) \cdot \D \head_{i-1}(x) \cdot v$ from Lemma \ref{lem:forward_prop} and $x_{i+1} = f_i(x_i)$.
\end{proof} 
\end{theorem}

Theorem \ref{thm:tgt_backprop} provides a recursive update formula for $\left(\left(\D \head_{i-1}(x) \cdot v \right) \lefthook \D^2\tail_i(x_i)\right)^*$, which backpropagates the error through the tangent network via multiplication by $\D^*f_i(x_i)$ and adding another term. Recall that the map $\D^*\tail_{i+1}(x_{i+1})$ is calculated recursively using Theorem \ref{theorem:backprop}. Now, the main result for calculating $\nabla_{\theta_i} R(x; \theta)$ is presented. 

\begin{theorem} \label{thm:DF_righthook_star} For any $x, v \in E_1$ and $i \in \{1, \ldots, L\}$,
\begin{align*}
\left(\nabla_{\theta_i} \D F(x;\theta) \righthook v\right)^* &= \nabla^*_{\theta_i} f_i(x_i) \cdot \left((\D\head_i(x) \cdot v) \lefthook \D^2\tail_{i+1}(x_{i+1})\right)^* \\
& + \left((\D\head_{i-1}(x) \cdot v) \lefthook \D\nabla_{\theta_i} f_i(x_i)\right)^* \cdot \D^*\tail_{i+1}(x_{i+1}),
\end{align*}
where $F(x; \theta) = f_L \circ \cdots \circ f_1(x)$, $\head_i$ is defined as in \eqref{eqn:head}, and $\tail_i$ is defined as in \eqref{eqn:tail}.

\begin{proof}
For any $i \in \{1, \ldots, L\}$ and $U_i \in H_i$, 
\begin{align*}
\left(\nabla_{\theta_i}\D F(x; \theta) \righthook v\right) \cdot U_i &= \nabla_{\theta_i}\D F(x; \theta) \cdot (U_i, v) \\
&= \D \left(\nabla_{\theta_i} F(x; \theta) \cdot U_i\right) \cdot v \\
&= \D\left(\D \tail_{i+1}(\head_i(x)) \cdot \nabla_{\theta_i} f_i(\head_{i-1}(x)) \cdot U_i\right) \cdot v \\
&= \D^2 \tail_{i+1}(x_{i+1}) \cdot \left(\D\head_i(x) \cdot v, \nabla_{\theta_i} f_i(x_i) \cdot U_i\right) \\
&\qquad + \D\tail_{i+1}(x_{i+1}) \cdot \D\nabla_{\theta_i} f_i(x_i) \cdot (\D\head_{i-1}(x)\cdot v, U_i) \\
&= \left( (\D\head_i(x) \cdot v) \lefthook \D^2\tail_{i+1}(x_{i+1}) \right) \cdot \nabla_{\theta_i} f_i(x_i) \cdot U_i\\
&\qquad + \D\tail_{i+1}(x_{i+1}) \cdot \left((\D\head_{i-1}(x)\cdot v) \lefthook \D \nabla_{\theta_i} f_i(x_i)\right) \cdot U_i.
\end{align*}
Since this holds for all $U_i \in H_i$,
\begin{align*}
\left(\nabla_{\theta_i}\D F(x; \theta) \righthook v\right)  &= \left( (\D\head_i(x) \cdot v) \lefthook \D^2\tail_{i+1}(x_{i+1}) \right) \cdot \nabla_{\theta_i} f_i(x_i)\\
&\qquad + \D\tail_{i+1}(x_{i+1}) \cdot \left((\D\head_{i-1}(x)\cdot v) \lefthook \D \nabla_{\theta_i} f_i(x_i)\right).
\end{align*}
Taking the adjoint of this proves the theorem.
by the reversing property of the adjoint.
\end{proof}
\end{theorem}

Algorithm \ref{alg:general_highorder_backprop} presents a single iteration of a gradient descent algorithm to minimize $\mathcal{J}_R$ directly over the parameter set $\theta = \{\theta_1, \ldots, \theta_L\}$. This formula extends linearly to a batch of updates over several data points. To extend this to $R$ defined with multiple $(v_x, \beta_x)$ pairs as in \eqref{eqn:R_mult_terms}, then there must be a set $V^j = \{v_1^j, \ldots, v_{L+1}^j\}$ calculated for each pair; in Algorithm \ref{alg:general_highorder_backprop}, only the one set $\{v_1, \ldots, v_{L+1}\}$ is calculated. 

\begin{algorithm}
\caption{One iteration of gradient descent for a higher-order loss function}
 \label{alg:general_highorder_backprop}
\begin{algorithmic}
\Function{Descent Iteration}{$x, v_x, \beta_x, y, \theta_1, \ldots, \theta_L, \eta, \mu$}
\State $x_1 \gets x$
\State $v_1 \gets v_x$ \Comment $v_i = \D\head_{i-1}(x) \cdot v_x$ and $\D\head_0(x) = \mbox{identity}$
\For {$i \in \{1, \ldots, L\}$} \Comment $x_{L+1} = F(x; \theta)$ and $v_{L+1} = \D F(x;\theta) \cdot v_x$ 
\State $x_{i+1} \gets f_i(x_i)$
\State $v_{i+1} \gets \D f_i(x_i) \cdot v_i$ \Comment Lemma \ref{lem:forward_prop}
\EndFor
\For {$i \in \{L, \ldots, 1\}$}
\State $\tilde{\theta_i} \gets \theta_i$ \Comment Store $\theta_i$ for updating $\theta_{i-1}$
\If {$i = L$}
\State $e_t \gets 0$ \Comment $e_t = \left( v_{i+1} \lefthook \D^2\tail_{i+1}(x_{i+1})\right)^* \cdot \left(v_{L+1} - \beta_x\right)$
\State $e_v \gets v_{L+1} - \beta_x$ \Comment $e_v = \D^*\tail_{i+1}(x_{i+1}) \cdot \left(v_{L+1} - \beta_x\right)$
\State $e_y \gets x_{L+1} - y$ \Comment $e_y = \D^*\tail_{i+1}(x_{i+1}) \cdot \left(x_{L+1} - y\right)$
\Else \Comment Calculate $\D^*f_{i+1}(x_{i+1})$ with $\tilde{\theta}_{i+1}$ in this block
\State $e_t \gets \D^*f_{i+1}(x_{i+1}) \cdot e_t + \left(v_{i+1} \lefthook \D^2 f_{i+1}(x_{i+1}) \right)^* \cdot e_v$ \Comment Thm. \ref{thm:tgt_backprop}; use old $e_v$
\State $e_v \gets \D^*f_{i+1}(x_{i+1}) \cdot e_v$ \Comment Thm. \ref{theorem:backprop}
\State $e_y \gets \D^*f_{i+1}(x_{i+1}) \cdot e_y$ \Comment Thm. \ref{theorem:backprop}
\EndIf
\State $\nabla_{\theta_i} J(x; \theta) \gets \nabla_{\theta_i}^* f_i(x_i) \cdot e_y$ \Comment Thms. \ref{theorem:nabla:F:general} and \ref{theorem:nabla_F}
\State $\nabla_{\theta_i} R(x; \theta) \gets \nabla_{\theta_i}^* f_i(x_i) \cdot e_t + \left(v_i \lefthook \D\nabla_{\theta_i}f_i(x_i)\right)^* \cdot e_v$ \Comment Thms. \ref{theorem:nabla:theta;R} and \ref{thm:DF_righthook_star}
\State $\theta_i \gets \theta_i - \eta (\nabla_{\theta_i} J(x;\theta) + \mu \nabla_{\theta_i} R(x; \theta))$
\EndFor
\EndFunction
\end{algorithmic}
\end{algorithm}

\section{Application 1: Standard Multilayer Perceptron} \label{sec:MLP}
The first network considered is a standard multilayer perceptron (MLP). The input data here is $x \in \R{n_1}$, and the output is $F \in \R{n_{L+1}}$ when the MLP is assumed to have $L$ layers. The single-layer function $f_i : \R{n_i} \times (\R{n_{i+1} \times n_i} \times \R{n_{i+1}}) \rightarrow \R{n_{i+1}}$ takes in the data at the $i^{th}$ layer --- $x_i \in \R{n_i}$ --- along with parameters $W_i \in \R{n_{i+1} \times n_i}$ and $b_i \in \R{n_{i+1}}$, and outputs the data at the $(i+1)^{th}$ layer, i.e. 
\[
x_{i+1} = f_i(x_i; W_i, b_i) \in \R{n_{i+1}}. 
\]
The dependence of $f_i$ on its parameters $(W_i, b_i)$ will often be suppressed throughout this section, i.e. $f_i(x_i; W_i, b_i) \equiv f_i(x_i)$, for convenience when composing functions.  It is assumed that every vector space used here is equipped with the usual Euclidean inner product. Thus, the inner product of  two matrices or vectors $A$ and $B$ of equal size is computed as
\[
\langle A, B \rangle = \tr (A^TB). 
\]
As a corollary, $\langle A, BC\rangle = \langle B^TA, C \rangle = \langle AC^T, B\rangle$
for any matrices or vectors $A$, $B$ and $C$ so that the inner product $\langle A, BC\rangle$ is valid. Every vector in each $\R{n_i}$ is treated as an $n_i \times 1$ matrix by default.

The explicit action of the layer-wise function $f_i$ can be described via an elementwise function $\bignl_i : \R{n_{i+1}} \rightarrow \R{n_{i+1}}$, with associated elementwise operation $\smallnl_i : \R{} \rightarrow \R{}$, as 
\begin{align} \label{eqn:MLP_f_i}
f_i(x_i) = \bignl_i (W_i \cdot x_i + b_i)
\end{align}
for any $x_i \in \R{n_i}$, where $\cdot$ denotes matrix-vector multiplication.  The elementwise function $\bignl_i : \R{n_{i+1}} \rightarrow \R{n_{i+1}}$ is defined as in \eqref{eqn:elem_fn}. The operation $\smallnl_i $ is nonlinear, so $\bignl_i$ is known as an elementwise \emph{nonlinear} function, or \emph{elementwise nonlinearity}. The derivative maps $\D\bignl_i$ and $\D^2\bignl_i$ can be calculated using Propositions \ref{prop:DS} and \ref{prop:DtwoS}, respectively. 

\begin{remark}
The maps $\D\bignl_i$ and $\D^2\bignl_i$ clearly depend on the choice of nonlinearity $\smallnl_i$. Some common choices and their derivatives are given in Table \ref{tab:nonlinearities}. Note that $H$ is the Heaviside step function, and $\sinh$ and $\cosh$ are the hyperbolic sine and cosine functions, respectively. Table \ref{tab:nonlinearities} is not a complete description of all possible nonlinearities. 
\begin{table}[ht]
\centering
\caption{Common nonlinearities, along with their first and second derivatives}
\label{tab:nonlinearities}
\begin{tabular}{|c|c|c|c|}
\hline
\textbf{Name}      & \textbf{Definition} & \textbf{First Derivative} & \textbf{Second Derivative} \\ \hline \T\B
tanh &$\smallnl_i(x) \coloneqq \frac{\sinh(x)}{\cosh(x)}$ & $\smallnl_i'(x) = \frac{4 \cosh^2(x)}{(\cosh(2x) + 1)^2} $ & $\smallnl_i''(x) = -\frac{8 \sinh(2x) \cosh^2(x)}{(\cosh(2x) + 1)^3}$                          \\ \hline \T\ 
Sigmoidal          & $\smallnl_i(x) \coloneqq \frac{1}{1 + \exp(-x)}$& $\smallnl'_i(x) = \smallnl_i(x) \left(1 - \smallnl_i(x)\right)$                          &   $\smallnl''_i(x) = \smallnl'_i(x) \left(1 - 2 \smallnl_i(x)\right)$                         \\ \hline \T\B
Ramp      &  $\smallnl_i(x) \coloneqq \max(0, x)$ & $\smallnl'_i(x) = H(x)$        & $\smallnl''_i(x) = 0$                            \\ \hline
\end{tabular}
\end{table}
\end{remark}

\subsection{Gradient Descent for Standard Loss Function} \label{sec:first_order_MLP}
Consider the loss function $J$ given in  \eqref{eqn:standard_loss}. Its gradient  with respect to the parameters $W_i$ and $b_i$ can now be calculated separately at each layer $i \in \{1, \ldots, L\}$, since $W_i$ and $b_i$ are both independent of each other and independent of other layers $j \neq i$. First, the derivatives of $f_i$ and their adjoints are computed:

\begin{lemma}\label{lemma:Dfs:MLP}
Consider the function $f_i$ defined in \eqref{eqn:MLP_f_i}. Then,  for any $x_i \in \R{n_i}$ and any $U_i \in \R{n_{i+1} \times n_{i}}$
\begin{align}
\nabla_{W_i} f_i(x_i) \cdot U_i 
&= \D\bignl_i(z_i) \cdot U_i \cdot x_i, \label{nabla:W:f:MLP:lemma}\\
\nabla_{b_i}f_i(x_i) &= \D\bignl_i(z_i), \label{nabla:b:f:MLP:lemma}
\end{align}
where $z_i = W_i \cdot x_i + b_i$, and
\begin{equation}\label{Dfi:MLP:lemma}
\D f_i(x_i) =  \D\bignl_i(z_i) \cdot W_i.
\end{equation}
This holds for any $i \in \{1, \ldots, L\}$.
\begin{proof}
For any $U_i \in \R{n_{i+1} \times n_i}$, 
\begin{align*}
\nabla_{W_i} f_i(x_i) \cdot U_i &= \left . \dd{}{t}\left(\bignl_i((W_i+tU_i) \cdot x_i + b_i)\right)\right |_{t=0} = \D\bignl_i(W_i \cdot x_i + b_i) \cdot U_i \cdot x_i,
\end{align*}
which proves \eqref{nabla:W:f:MLP:lemma}. The other equations can be proven similarly.
\end{proof}
\end{lemma}

\begin{lemma} \label{lem:D_star_f_MLP}
For any $i \in \{1, \ldots, L\}$, $x_i \in \R{n_i}$ and $u \in \R{n_{i+1}}$, 
\begin{align}
\nabla^*_{W_i} f_i(x_i) \cdot u  &= \left(\bignl_i'(z_i) \odot u \right) x_i^T,\label{nabla:star:Wi:f:MLP}\\
\nabla^*_{b_i}f_i(x_i) &= \D\bignl_i(z_i), \label{nabla:star:bi:f:MLP}
\end{align}
where $z_i = W_i \cdot x_i + b_i$, and
\begin{equation}\label{Dstar;f:MLP}
\D^*f_i(x_i) = W_i^T \cdot \D\bignl_i(z_i).
\end{equation}
\begin{proof}
By Lemma \ref{lemma:Dfs:MLP}, for any $u \in \R{n_{i+1}}$ and any $U_i \in \R{n_{i+1} \times n_i}$
\begin{align*}
\ip{u}{\nabla_{W_i} f_i(x_i) \cdot U_i} &= \ip{z}{\D\bignl_i(z_i) \cdot U_i \cdot x_i} \\
&= \ip{\D\bignl_i(z_i) \cdot u}{U_i \cdot x_i} \\
&= \ip{\left(\D\bignl_i(z_i) \cdot u\right) x_i^T}{U_i},
\end{align*}
which implies
\begin{equation} \label{eqn:nabla_star_Wi_f_MLP_temp}
\nabla^*_{W_i} f_i(x_i) \cdot u = \left(\D\bignl_i(z_i) \cdot u\right) x_i^T = \left(\bignl_i'(z_i) \odot u\right) x_i^T,
\end{equation}
which proves \eqref{nabla:star:Wi:f:MLP}. Equations \eqref{nabla:star:bi:f:MLP} and \eqref{Dstar;f:MLP} follow from taking the adjoints of \eqref{nabla:b:f:MLP:lemma} and \eqref{Dfi:MLP:lemma} and using the self-adjointness of $\D \bignl_i(z_i)$. 
\end{proof}
\end{lemma}

The next result demonstrates how to backpropagate the error in the network.
\begin{theorem}[Backpropagation in MLP] \label{thm:MLP_backprop} For $f_i$ defined as in \eqref{eqn:MLP_f_i} and $\tail_i$ as defined in \eqref{eqn:tail},
\begin{align} \label{eqn:MLP_D_tail}
\D\tail_i(x_i) = \D\tail_{i+1}(x_{i+1}) \cdot \D \bignl_i(z_i) \cdot W_i,
\end{align}
where $x_{i+1} = f_i(x_i)$ for all $i \in \{1, \ldots, L\}$, $z_i = W_i \cdot x_i + b_i$, and $\tail_{L+1}$ is the identity. Furthermore, for any $u \in \R{n_{L+1}}$,
\begin{equation}\label{eqn:MLP_Dstar_tail}
\D^*\tail_i(x_i) \cdot u = W_i^T \cdot \left(\bignl_i'(z_i) \odot (\D^*\tail_{i+1}(x_{i+1}) \cdot u)\right).
\end{equation}

\begin{proof}
Pick any $v \in \R{n_i}$. By the Theorem \ref{theorem:backprop} and Lemma \ref{lemma:Dfs:MLP},
\begin{align*}
\D \tail_i(v) &= \D \tail_{i+1}(f_i(v)) \cdot \D f_i(v) \\
&= \D \tail_{i+1}(f_i(v)) \cdot \D\bignl_i(W_i \cdot v + b_i) \cdot W_i
\end{align*}
Then, setting $v = x_i$, equation \eqref{eqn:MLP_D_tail} is proven since $x_{i+1} = f_i(x_i)$. 

Now, by taking the adjoint of the above equation
\[
\D^*\tail_i(x_i) = W_i^* \cdot \D^* \bignl_i(z_i) \cdot \D^*\tail_{i+1}(x_{i+1}),
\]
where $W_i^* = W_i^T$.  Also, $\D^*\bignl_i(z_i) = \D\bignl_i(z_i)$ from Proposition \ref{prop:DS}. Thus, applying $\D^*\tail_i(x_i)$ to any $v \in \R{n_{L+1}}$ gives
\begin{align*}
\D^*\tail_i(x_i) \cdot v &=  W_i^T \cdot \D\bignl_i(z_i) \cdot \D^*\tail_{i+1}(x_{i+1}) \cdot v \\
&= W_i^T \cdot \left(\bignl_i'(z_i) \odot (\D^*\tail_{i+1}(x_{i+1}) \cdot v)\right).
\end{align*}
This is true for any $i \in \{1, \ldots, L\}$, so the proof is complete.
\end{proof}
\end{theorem}

The above theorem demonstrates how to calculate $\D^*\tail_i(x_i)$ recursively, which is needed to backpropagate the error throughout the network. This will be necessary to compute the main MLP result presented in the next theorem. 

\begin{theorem} 
Let $J$ be defined as in \eqref{eqn:standard_loss}, $\theta = \{W_1, \ldots, W_L, b_1, \ldots, b_L\}$ represent the parameters, $x \in \R{n_1}$ be an input with associated known output $y \in \R{n_{L+1}}$, and $F(x; \theta)$ be defined as in \eqref{eqn:F}. Then, the following equations hold for any $i \in \{1, \ldots, L\}$: 
\begin{align}
\nabla_{W_i}J(x; \theta) &= \left(\bignl_i'\left(z_i\right) \odot \left(\D^*\tail_{i+1}(x_{i+1}) \cdot e\right)\right) x_i^T , \label{nabla:W:J:MLP}\\
\nabla_{b_i} J(x; \theta) &= \bignl'_i(z_i) \odot \left(\D^*\tail_{i+1}(x_{i+1}) \cdot e \right), \label{nabla:b:J:MLP}
\end{align}
where $x_i = \head_{i-1}(x)$, $z_i = W_i \cdot x_i + b_i$, and the prediction error $e$ is given by
\[
e = F(x; \theta) - y \in \R{n_{L+1}}.
\]
\begin{proof} 
By Theorem \ref{theorem:nabla:F:general}
\begin{align}
\nabla_{W_i} J(x; \theta) &= \nabla^*_{W_i} F(x; \theta) \cdot e,  \label{nabla:Wi:J:MLP}\\
\nabla_{b_i} J(x; \theta) &= \nabla^*_{b_i} F(x; \theta) \cdot e. \label{nabla:bi:J:MLP}
\end{align}
From Theorem \ref{theorem:nabla_F},
\begin{align}
\nabla^*_{W_i} F(x; \theta) \cdot e &= \nabla^*_{W_i} f_i(x_i) \cdot \D^*\tail_{i+1}(x_{i+1}) \cdot e, \label{nabla:Wi:star:F:MLP} \\
\nabla^*_{b_i} F(x; \theta) \cdot e &= \nabla^*_{b_i} f_i(x_i) \cdot \D^*\tail_{i+1}(x_{i+1}) \cdot e.\label{nabla:bi:star:F:MLP}
\end{align}
Recall that $\D^*\tail_{i+1}(x_{i+1}) \cdot e$ is calculated recursively via Theorem \ref{thm:MLP_backprop}.
Then, \eqref{nabla:W:J:MLP} follows from \eqref{nabla:Wi:J:MLP}, \eqref{nabla:Wi:star:F:MLP} and \eqref{nabla:star:Wi:f:MLP}, i.e.
\begin{align*}
\nabla_{W_i} J(x; \theta) &= \nabla^*_{W_i} F(x;\theta) \cdot e \\
&= \nabla^*_{W_i} f_i(x_i) \cdot \D^*\tail_{i+1}(x_{i+1}) \cdot e \\
&= \left(\bignl_i'(z_i) \odot \left(\D^*\tail_{i+1}(x_{i+1}) \cdot e\right)\right) x_i^T.
\end{align*}
Similarly, \eqref{nabla:b:J:MLP} follows from \eqref{nabla:bi:J:MLP}, \eqref{nabla:Wi:star:F:MLP} and  \eqref{nabla:star:bi:f:MLP}, i.e.
\begin{align*}
\nabla_{b_i}J(x;\theta) &= \nabla^*_{b_i}F(x;\theta) \cdot e \\
&= \nabla^*_{b_i}f_i(x_i) \cdot \D^*\tail_{i+1}(x_{i+1}) \cdot e \\
&=  \bignl'_i(z_i) \odot \left(\D^*\tail_{i+1}(x_{i+1}) \cdot e \right).
\end{align*}
This completes the proof, which is valid for all $i \in \{1, \ldots, L\}$.
\end{proof}
\end{theorem}

Given the above results, a gradient descent algorithm can be developed to minimize $J$ with respect to each $W_i$ and $b_i$, for a given data point $x$ and \emph{learning rate} $\eta$. One iteration of this is given in Algorithm \ref{alg:MLP_grad_desc_1}. The output of the algorithm is an updated version of $W_i$ and $b_i$. This process can be extended additively to a batch of updates by summing the individual contributions of each $x$ to the gradient of $J(x; \theta)$. 

\begin{algorithm}
\caption{One iteration of gradient descent in MLP}
 \label{alg:MLP_grad_desc_1}
\begin{algorithmic}
\Function{Descent Iteration}{$x, y, W_1, \ldots, W_L, b_1, \ldots, b_L, \eta$}
\State $x_1 \gets x$
\For {$i \in \{1, \ldots, L\}$} \Comment $x_{L+1} = F(x; \theta)$ 
\State $z_i \gets W_i \cdot x_i + b_i$ 
\State $x_{i+1} \gets \bignl_i(z_i)$
\EndFor
\For {$i \in \{L, \ldots, 1\}$}
\State $\tilde{W_i} \gets W_i$ \Comment Store old $W_i$ for updating $W_{i-1}$
\If {$i = L$} \Comment $e = \D^*\tail_{i+1}(x_{i+1}) \cdot \left(x_{L+1} - y\right)$
\State $e \gets x_{L+1} - y$   \Comment $\tail_{L+1} = \textup{identity}$ 
\Else
\State $e \gets \tilde{W}_{i+1}^T \cdot (\bignl'_{i+1}(z_{i+1}) \odot e)$ \Comment \eqref{eqn:MLP_Dstar_tail}  
\EndIf
\State $\nabla_{b_i} J(x; \theta) \gets \bignl_i'(z_i) \odot e$  \Comment \eqref{nabla:b:J:MLP} 
\State $\nabla_{W_i} J(x; \theta) \gets \left(\bignl_i'(z_i) \odot e\right) x_i^T$ \Comment \eqref{nabla:W:J:MLP}
\State $b_i \gets b_i - \eta \nabla_{b_i} J(x;\theta)$
\State $W_i \gets W_i - \eta \nabla_{W_i} J(x;\theta)$
\EndFor
\EndFunction
\end{algorithmic}
\end{algorithm}

\subsection{Gradient Descent for Higher-Order Loss Function}
The goal now is to perform a gradient descent iteration for a higher-order loss function of the form \eqref{eqn:curly_J}. Since the gradients of $J$ are already understood, it is only necessary to compute the gradients of $R$, defined in \eqref{eqn:R}, with respect to $W_i$ and $b_i$ for all $i \in \{1, \ldots, L\}$. This will involve forward and backward propagation through the tangent network, and then the calculation of $\left(\nabla \D F(x; \theta) \righthook v\right)^*$, as in Theorem \ref{thm:DF_righthook_star}. First, relevant single-layer derivatives will be presented as in the previous section. 

\begin{lemma}
Consider the function $f_i$ defined in \eqref{eqn:MLP_f_i}. Then, for any $x_i, v \in \R{n_i}$ and $U_i \in \R{n_{i+1} \times n_i}$,
\begin{align}
\left(v \lefthook \D\nabla_{W_i} f_i(x_i)\right) \cdot U_i &= \D^2 \bignl_i(z_i) \cdot (W_i \cdot v, U_i \cdot x_i) + \D\bignl_i(z_i) \cdot U_i \cdot v \label{eqn:v_lh_D_W_f_MLP} \\
\left(v \lefthook \D\nabla_{b_i} f_i(x_i)\right) &= \left((W_i \cdot v) \lefthook \D^2\bignl_i(z_i)\right) \label{eqn:v_lh_D_b_MLP} \\ 
\left(v \lefthook \D^2 f_i(x_i)\right) &= \left( (W_i\cdot v) \lefthook \D^2\bignl_i(z_i)\right) \cdot W_i, \label{eqn:v_lh_D2_f_MLP}
\end{align}
where $z_i = W_i \cdot x_i + b_i$. Furthermore, for any $y \in \R{n_{i+1}}$, 
\begin{align}
\left(v \lefthook \D\nabla_{W_i} f_i(x_i)\right)^* \cdot y &= \left[\bignl_i''(z_i) \odot (W_i \cdot v) \odot y\right] x_i^T + \left[\bignl_i'(z_i) \odot y\right] v^T \label{eqn:v_lh_D_S_W_MLP},\\
\left(v \lefthook \D\nabla_{b_i}f_i(x_i)\right)^* &= \left( (W_i \cdot v) \lefthook \D^2\bignl_i(z_i) \right), \label{eqn:v_lh_D_S_b_MLP}\\
\left(v \lefthook \D^2 f_i(x_i) \right)^* &= W_i^T \cdot \left( (W_i \cdot v) \lefthook \D^2\bignl_i(z_i)\right). \label{eqn:v_lh_D2_S_f_MLP}
\end{align}

\begin{proof}
First, equation \eqref{eqn:v_lh_D_W_f_MLP} is proven directly:
\begin{align*}
\left(v \lefthook \D\nabla_{W_i} f_i(x_i)\right) \cdot U_i &= \D\left(\nabla_{W_i} f_i(x_i) \cdot U_i\right) \cdot v \\
&= \D \left( \D\bignl_i(z_i) \cdot U_i \cdot x_i\right) \cdot v \\
&= \D^2 \bignl_i(z_i) \cdot (W_i \cdot v, U_i \cdot x) + \D\bignl_i(z_i) \cdot U_i \cdot v,
\end{align*}
where the second line comes from \eqref{nabla:W:f:MLP:lemma} and the last line follows from Lemma \ref{lem:chain_rule_2}. Equations \eqref{eqn:v_lh_D_b_MLP} and \eqref{eqn:v_lh_D2_f_MLP} can be proven similarly. 

Next, equation \eqref{eqn:v_lh_D_S_W_MLP} is proven directly. For any $y \in \R{n_{i+1}}$, 
\begin{align*}
\ip{y}{\left(v \lefthook \D\nabla_{W_i} f_i(x_i)\right) \cdot U_i} &= \ip{y}{\left((W_i \cdot v) \lefthook \D^2\bignl_i(z_i)\right) \cdot U_i \cdot x_i + \D\bignl_i(x_i) \cdot U_i \cdot v} \\
&= \left\langle \left[\left( (W_i \cdot v) \lefthook \D^2\bignl_i(z_i) \right) \cdot y\right] x_i^T + \left[ \D\bignl_i(z_i) \cdot y \right] v^T, U_i \right\rangle,
\end{align*}
since $\D\bignl_i(z_i)$ and $\left(v \lefthook \D^2\bignl_i(z_i)\right)$ are both self-adjoint. Since this is true for any $U_i$,
\begin{align} \label{eqn:v_lh_D_S_W_MLP_temp}
\left(v \lefthook \D\nabla_{W_i} f_i(x_i)\right)^* \cdot y =  \left[\left( (W_i \cdot v) \lefthook \D^2\bignl_i(z_i) \right) \cdot y\right] x_i^T + \left[ \D\bignl_i(z_i) \cdot y \right] v^T,
\end{align}
which is equation \eqref{eqn:v_lh_D_S_W_MLP} once the definitions of $\D\bignl_i$ and $\D^2\bignl_i$ are substituted in. 

Equations \eqref{eqn:v_lh_D_S_b_MLP} and \eqref{eqn:v_lh_D2_S_f_MLP} are direct consequences of \eqref{eqn:v_lh_D_b_MLP} and \eqref{eqn:v_lh_D2_f_MLP}, respectively, using the reversing property of the adjoint and the self-adjointness of $\D\bignl_i(z_i)$ and $\left(v \lefthook \D^2\bignl_i(z_i)\right)$. 
\end{proof}
\end{lemma}

\begin{theorem} \label{theorem:FP:alpha:high:MLP}
For $f_i$ defined as in \eqref{eqn:MLP_f_i}, $\head_i$ defined as in \eqref{eqn:head}, and $x, v \in \R{n_1}$, 
\[
\D\head_i(x) \cdot v = \bignl'_i(z_i) \odot (W_i \cdot \D\head_{i-1}(x) \cdot v),
\]
where $x_i = \head_{i-1}(x)$, $z_i = W_i \cdot x_i + b_i$, and $i \in \{1, \ldots, L\}$. Also, $\D \head_0(x) \cdot v = v.$
\begin{proof} For all $i \in \{1, \ldots, L\}$, by Lemma \ref{lem:forward_prop}, Proposition \ref{prop:DS} and equation \eqref{Dfi:MLP:lemma},
\begin{align*}
\D\head_i(x) \cdot v &= \D f_i(\head_{i-1}(x)) \cdot \D\head_{i-1}(x) \cdot v \\
&= \bignl'_i(z_i) \odot (W_i \cdot \D\head_{i-1}(x) \cdot v),
\end{align*}
where $\head_0(x) = x$ and $x_i = \head_{i-1}(x)$. Furthermore, $\D\head_0(x)$ is the identity since $\head_0$ is the identity. 
\end{proof}
\end{theorem}
This is an explicit representation of the forward propagation through the tangent network. The next lemma describes the backpropagation through the tangent network.

\begin{theorem}[Tangent Backpropagation in MLP]\label{theorem:tangent:BP:MLP}
Let $\head_i$ and $\tail_i$ be defined as in \eqref{eqn:head} and \eqref{eqn:tail}, respectively. Let $f_i$ be defined as in \eqref{eqn:MLP_f_i}. Then, for any $i \in \{1, \ldots, L\}$, $x, v_1 \in \R{n_1}$, and $v_2 \in \R{n_{L+1}}$, 
\begin{align*}
\left( (\D \head_{i-1}(x) \cdot v_1) \lefthook \D^2\tail_i(x_i)\right)^*\! \cdot\! v_2 &= W_i^T \cdot \left\{\bignl_i'(z_i) \odot \left[ \left((\D\head_i(x) \cdot v_1) \lefthook \D^2\tail_{i+1}(x_{i+1})\right)^* \cdot v_2\right] \right\} \\
&+ W_i^T\! \cdot\! \left\{ \bignl''_i(z_i) \odot (W_i \!\cdot\! \D\head_{i-1}(x) \cdot v_1) \odot (\D^*\tail_{i+1}(x_{i+1}) \!\cdot\! v_2) \right\},
\end{align*}
where $x_i = \head_{i-1}(x)$ and $z_i = W_i \cdot x_i + b_i$. Also, 
\[
\left( (\D\head_L(x) \cdot v_1) \lefthook \D^2 \tail_{L+1}(x_{L+1})\right)^* \cdot v_2 = 0.
\]

\begin{proof}
Theorem \ref{thm:tgt_backprop} states that for any $i \in \{1, \ldots, L\}$, 
\begin{align} \label{eqn:ref_tgt_backprop}
\left(\left(\D \head_{i-1}(x) \cdot v_1 \right) \lefthook \D^2\tail_i(x_i)\right)^* \cdot v_2 &= \D^*f_i(x_i) \cdot \left((\D \head_i(x) \cdot v_1) \lefthook \D^2\tail_{i+1}(x_{i+1}) \right)^* \cdot v_2 \nonumber \\
&+ \left(\left(\D \head_{i-1}(x)\! \cdot\! v_1 \right) \lefthook \D^2 f_i(x_i)\right)^* \cdot \D^*\tail_{i+1}(x_{i+1}) \cdot v_2.
\end{align}
By \eqref{Dstar;f:MLP}, 
\[
\D^*f_i(x_i) = W_i^T \cdot \D\bignl_i(z_i).
\]
Furthermore, by \eqref{eqn:v_lh_D2_S_f_MLP},
\[
\left((\D\head_{i-1}(x) \cdot v_1) \lefthook \D^2f_i(x_i)\right)^* = W_i^T \cdot \left((W_i \cdot \D\head_{i-1}(x) \cdot v_1) \lefthook \D^2\bignl_i(z_i)\right).
\]
These results can be substituted into equation \eqref{eqn:ref_tgt_backprop} to obtain the final result:
\begin{align*}
\left(\left(\D \head_{i-1}(x) \cdot v_1 \right) \lefthook \D^2\tail_i(x_i)\right)^* \!\cdot\! v_2 &= W_i^T \cdot \D\bignl_i(z_i) \cdot \left((\D \head_i(x) \cdot v_1) \lefthook \D^2\tail_{i+1}(x_{i+1}) \right)^* \cdot v_2 \\
&+  W_i^T \cdot \left((W_i \cdot \D\head_{i-1}(x) \cdot v_1) \lefthook \D^2\bignl_i(z_i)\right) \cdot \D^*\tail_{i+1}(x_{i+1}) \cdot v_2 \\
&= W_i^T \cdot \left\{\bignl_i'(z_i) \odot \left[ \left((\D\head_i(x) \cdot v_1) \lefthook \D^2\tail_{i+1}(x_{i+1})\right)^* \cdot v_2\right] \right\} \\
&+ W_i^T\! \cdot\! \left\{ \bignl''_i(z_i) \odot (W_i\! \cdot\! \D\head_{i-1}(x)\! \cdot \! v_1) \odot (\D^*\tail_{i+1}(x_{i+1}) \cdot v_2) \right\}.
\end{align*}
This is true even for $i = 1$ since $\head_0$ is the identity. For $i = L+1$, $\tail_{L+1}$ is also the identity, so $((\head_L(x) \cdot v_1) \lefthook \D^2\tail_{L+1}(x_{L+1}))^*$ is the zero operator. Thus, the result is proven. 
\end{proof}
\end{theorem}

Note that the first term in  the tangent backpropagation expression in Theorem \ref{theorem:tangent:BP:MLP} is the recursive part, and the second term can be calculated at each stage once $\D^*\tail_{i+1}(x_{i+1})$ is calculated. The maps $(\nabla_{W_i} \D F(x;\theta) \righthook v)^*$ and $(\nabla_{b_i} \D F(x;\theta) \righthook v)^*$ are calculated in the next theorem as the final step in the gradient descent puzzle. 

\begin{theorem}\label{theorem:ultimate:BP:h:MLP} 
Let $v \in \R{n_1}$ and $e \in \R{n_{L+1}}$. Then, with $\head_i$ and $\tail_i$ defined as in \eqref{eqn:head} and \eqref{eqn:tail}, respectively, and $x_i = \head_{i-1}(x)$,
\begin{align}
\left(\nabla_{W_i} \D F(x;\theta) \righthook v\right)^* \cdot e &= \left\{\bignl_i'(z_i) \odot \left[\left((\D \head_i(x)\cdot v) \lefthook \D^2\tail_{i+1}(x_{i+1})\right)^* \cdot e\right]\right\} x_i^T \nonumber \\
&\qquad + \left(\bignl''_i(z_i) \odot (W_i \cdot \D\head_{i-1}(x) \cdot v) \odot (\D^*\tail_{i+1}(x_{i+1}) \cdot e) \right)x_i^T \label{eqn:nabla_W_D_F_MLP} \\
&\qquad + \left(\bignl_i'(z_i) \odot (\D^*\tail_{i+1}(x_{i+1}) \cdot e)\right) (\D\head_{i-1}(x)\cdot v)^T, \nonumber\\
\left(\nabla_{b_i} \D F(x;\theta) \righthook v\right)^* \cdot e &= \bignl_i'(z_i) \odot \left[\left((\D \head_i(x)\cdot v) \lefthook \D^2\tail_{i+1}(x_{i+1})\right)^* \cdot e \right] \nonumber \\
&\qquad + \bignl_i''(z_i) \odot (W_i \cdot \D\head_{i-1}(x)\cdot v) \odot \left[\D^*\tail_{i+1}(x_{i+1}) \cdot e \right], \label{eqn:nabla_b_D_F_MLP}
\end{align}
for any $i \in \{1, \ldots, L\}$, where $z_i = W_i \cdot x_i + b_i$. 

\begin{proof}
Theorem \ref{thm:DF_righthook_star} says that for any $e \in \R{n_{L+1}}$, 
\begin{align}
\left(\nabla_{\theta_i} \D F(x;\theta) \righthook v\right)^* \cdot e &= \nabla^*_{\theta_i} f_i(x_i) \cdot \left((\D\head_i(x) \cdot v) \lefthook \D^2\tail_{i+1}(x_{i+1})\right)^* \cdot e \nonumber \\
& + \left((\D\head_{i-1}(x) \cdot v) \lefthook \D\nabla_{\theta_i} f_i(x_i)\right)^* \cdot \D^* \tail_{i+1}(x_{i+1}) \cdot e,  \label{eqn:nabla_D_F_MLP}
\end{align}
where $\theta_i$ is a generic parameter at layer $i$, for $i \in \{1, \ldots, L\}$. 

When $\theta_i = W_i$, equations \eqref{eqn:nabla_star_Wi_f_MLP_temp} and \eqref{eqn:v_lh_D_S_W_MLP_temp} can be substituted into \eqref{eqn:nabla_D_F_MLP} to obtain
\begin{align} 
\left(\nabla_{W_i} \D F(x;\theta) \righthook v \right)^* \cdot e &= \left\{\D\bignl_i(z_i) \cdot \left( \left(\D\head_i(x) \cdot v\right) \lefthook \D^2\tail_{i+1}(x_{i+1}) \right)^* \cdot e \right\} x_i^T \nonumber \\
&\quad + \left\{ \left( \left(W_i \cdot \D\head_{i-1}(x) \cdot v\right) \lefthook \D^2\bignl_i(z_i) \right) \cdot \D^*\tail_{i+1}(x_{i+1}) \cdot e\right\} x_i^T \label{eqn:nabla_W_D_F_MLP_temp} \\
&\quad + \left( \D\bignl_i(z_i) \cdot \D^*\tail_{i+1}(x_{i+1}) \cdot e\right) \left(\D\head_{i-1}(x) \cdot v\right)^T. \nonumber
\end{align}
Equation \eqref{eqn:nabla_W_D_F_MLP} is then obtained upon substituting the expressions for $\D\bignl_i(z_i)$ and $\D^2\bignl_i(z_i)$ into \eqref{eqn:nabla_W_D_F_MLP_temp}. 

Similarly, when $\theta_i = b_i$, equations \eqref{nabla:star:bi:f:MLP} and \eqref{eqn:v_lh_D_S_b_MLP} can be substituted into \eqref{eqn:nabla_D_F_MLP} to obtain
\begin{align}
\left(\nabla_{b_i} \D F(x;\theta) \righthook v \right)^* \cdot e &= \D\bignl_i(z_i) \cdot \left( \left(\D\head_i(x) \cdot v\right) \lefthook \D^2 \tail_{i+1}(x_{i+1})\right)^* \cdot e \nonumber \\
&\quad + \left( \left(W_i \cdot \D\head_{i-1}(x) \cdot v\right) \lefthook \D^2\bignl_i(z_i) \right) \cdot \D^*\tail_{i+1}(x_{i+1}) \cdot e. \label{eqn:nabla_b_D_F_MLP_temp}
\end{align}
As before, equation \eqref{eqn:nabla_b_D_F_MLP} is obtained by substituting the expressions for $\D\bignl_i(z_i)$ and $\D^2\bignl_i(z_i)$ into \eqref{eqn:nabla_b_D_F_MLP_temp}. 
\end{proof}
\end{theorem}

From Theorem \ref{theorem:nabla:theta;R}, for $v_x \in \R{n_1}$ and $\beta_x \in \R{n_{L+1}}$ 
\begin{align*}
\nabla_{\theta_i}R(x; \theta) = \left(\nabla_{\theta_i} \D F(x; \theta) \righthook v_x\right)^* \cdot \left(\D F(x; \theta) \cdot v_x - \beta_x\right),
\end{align*}
for $\theta_i$ equal to one of $W_i$ or $b_i$. Substitute $v= v_x$ and $e = (\D F(x; \theta) \cdot v_x - \beta_x )$ in the formulas in Theorem \ref{theorem:ultimate:BP:h:MLP} to compute $\nabla_{W_i}R(x; \theta)$ and $\nabla_{b_i}R(x; \theta)$.   
Thus, one iteration of a gradient descent algorithm to minimize $\mathcal{J}_R = J + \mu R$ can now be given, since $\nabla_{\theta_i} J(x; \theta)$ and $\nabla_{\theta_i} R(x; \theta)$ can both be calculated. This is described in Algorithm \ref{alg:MLP_grad_desc_2}.  

\begin{algorithm}
\caption{One iteration of gradient descent for higher-order loss in MLP}
\label{alg:MLP_grad_desc_2}
\begin{algorithmic}
\Function{Descent Iteration}{$x, v_x, \beta_x, y, W_1, \ldots, W_L, b_1, \ldots, b_L, \eta, \mu$}
\State $x_1 \gets x$
\State $v_1 \gets v_x$ \Comment $v_i = \D\head_{i-1}(x) \cdot v_x$ and $\D\head_0(x) = $ identity 
\For {$i \in \{1, \ldots, L\}$} \Comment $x_{L+1} = F(x; \theta)$ and $v_{L+1} = \D F(x; \theta) \cdot v_x$ 
\State $z_i \gets W_i \cdot x_i + b_i$
\State $x_{i+1} \gets \bignl_i(z_i)$
\State $v_{i+1} \gets \bignl_i'(z_i) \odot (W_i \cdot v_i)$ \Comment Theorem \ref{theorem:FP:alpha:high:MLP}
\EndFor
\For {$i \in \{L, \ldots, 1\}$}
\State $\tilde{W_i} \gets W_i$ \Comment Store $W_i$ for updating $W_{i-1}$
\If {$i = L$}
\State $e_t \gets 0$ \Comment $e_t = \left(v_{i+1} \lefthook \D^2\tail_{i+1}(x_{i+1}) \right)^* \cdot \left(v_{L+1} - \beta_x\right)$
\State $e_v \gets v_{L+1} - \beta_x$ \Comment $e_v = \D^*\tail_{i+1}(x_{i+1}) \cdot \left(v_{L+1} - \beta_x\right)$ 
\State $e_y \gets x_{L+1} - y$ \Comment $e_y = \D^*\tail_{i+1}(x_{i+1}) \cdot \left(x_{L+1} - y\right)$ 
\Else
\State $e_t \gets \tilde{W}_{i+1}^T \cdot \left(\bignl_{i+1}'(z_{i+1}) \odot e_t + \bignl_{i+1}''(z_{i+1}) \odot (\tilde{W}_{i+1} \cdot v_{i+1}) \odot e_v\right)$ \Comment Theorem \ref{theorem:tangent:BP:MLP}
\State $e_v \gets \tilde{W}_{i+1}^T \cdot (\bignl'_{i+1}(z_{i+1}) \odot e_v)$ \Comment \eqref{eqn:MLP_Dstar_tail};  Update $e_v$ after update of $e_t$
\State $e_y \gets \tilde{W}_{i+1}^T \cdot (\bignl'_{i+1}(z_{i+1}) \odot e_y)$ \Comment \eqref{eqn:MLP_Dstar_tail}
\EndIf
\State $\nabla_{b_i} J(x; \theta) \gets \bignl_i'(z_i) \odot e_y$ \Comment \eqref{nabla:b:J:MLP} 
\State $\nabla_{W_i} J(x; \theta) \gets \left(\bignl_i'(z_i) \odot e_y\right) x_i^T$ \Comment \eqref{nabla:W:J:MLP}
\State $\nabla_{b_i} R(x; \theta) \gets \bignl_i'(z_i) \odot e_t + \bignl_i''(z_i) \odot (W_i \cdot v_i) \odot e_v$ \Comment Thm. \ref{theorem:ultimate:BP:h:MLP} for this and next line 
\State $\nabla_{W_i} R(x; \theta) \gets (\bignl_i'(z_i) \odot e_t + \bignl_i''(z_i) \odot (W_i \cdot v_i) \odot e_v) x_i^T + (\bignl_i'(z_i) \odot e_v) v_i^T$
\State $W_i \gets W_i - \eta (\nabla_{W_i} J(x;\theta) + \mu \nabla_{W_i} R(x; \theta))$
\State $b_i \gets b_i - \eta (\nabla_{b_i} J(x;\theta) + \mu \nabla_{b_i} R(x; \theta) )$
\EndFor
\EndFunction
\end{algorithmic}
\end{algorithm}

\section{Application 2: Deep Autoencoder} \label{sec:AE}
Now, a $2L$-layer autoencoder (AE) of the form given in Murphy, Chapter 28 \cite{murphy2012machine} is described in the framework of Section \ref{sec:main_formulation}. The layerwise function $f_i$ is slightly more complicated in this case because there is weight-sharing between differernt layers of the network. Introduce a function $\ind : \{1, \ldots, 2L\} \rightarrow \{1, \ldots, 2L\}$ to aid in network representation, defined as follows:
\begin{align} \label{eqn:ind_perm}
\ind(i) = 2L - i + 1, \quad \forall i \in \{1, \ldots, 2L\}.
\end{align}
This function has the property that $(\ind \circ \ind) (i) = i$, for all $i$. Then, the layerwise function $f_i : \R{n_i} \times \left(\R{n_{i+1} \times n_i} \times \R{n_{i+1}}\right) \rightarrow \R{n_{i+1}}$ can be represented in the following manner:
\begin{align*}
f_i(x_i; W_i, b_i) &= \bignl_i(W_i \cdot x_i + b_i), \quad &i &\in \{1, \ldots, L\} \\
f_i(x_i; W_{\ind(i)}, b_i) &= \bignl_i\left(\aelin_i(W_{\ind(i)}) \cdot x_i + b_i\right), \quad  &i &\in \{L+1, \ldots, 2L\}, 
\end{align*}
where $x_i \in \R{n_i}$ is the input to the $i^{th}$ layer, $W_i \in \R{n_{i+1} \times n_i}$ is the weight matrix, $b_i \in \R{n_i}$ is the bias vector at layer $i$, $\bignl_i : \R{n_{i+1}} \rightarrow \R{n_{i+1}}$ is the elementwise nonlinearity with corresponding elementwise operation $\smallnl_i$, and $\aelin_i \in \lin{\R{n_{\ind(i)+1} \times n_{\ind(i)}}}{\R{n_{\ind(i)} \times n_{\ind(i)+1}}}$ governs how the weights are shared between layer $i$ and $\ind(i)$. The structure of the autoencoder is to encode for the first $L$ layers, and decode for the next $L$ layers, with the dimensions being preserved according to:
\[
n_{L + j} = n_{L - j + 2}, \quad \forall j \in \{2, \ldots, L + 1\}.
\] 
In \cite{murphy2012machine} and other similar examples, $\aelin_i$ is the matrix transpose operator at each layer, although it is kept general in this paper. However, for that particular case, the adjoint is calculated according to the following lemma.
\begin{lemma}
Let $\aelin \in \lin{\R{n \times m}}{\R{m \times n}}$ be defined as $\aelin(U) = U^T$ for all $U \in \R{n \times m}$. Then, 
\[
\aelin^*(W) = W^T 
\]
for all $W \in \R{m \times n}$.
\begin{proof}
For any $U \in \R{n \times m}$ and $W \in \R{m \times n}$, 
\[
\ip{W}{\aelin(U)} = \ip{W}{U^T} = \tr(W U) = \tr(U W) = \ip{U}{W^T},
\]
which proves the result by the symmetry of $\ip{\,}{}$. 
\end{proof}
\end{lemma} 

Now, introduce the following notation to represent the $f_i$ in a more compact manner:
\[
K_i = 
\begin{cases}
W_i, &1 \leq i \leq L \\
\aelin_i(W_{\ind(i)}), & L+1 \leq i \leq 2L
\end{cases}
\]
Then, the action of layer $i$ --- $f_i$ --- can be simply represented as
\begin{align} \label{eqn:AE_f_i}
f_i(x_i) = \bignl_i(K_i \cdot x_i + b_i), 
\end{align}
where the explicit dependence on the parameters $K_i$ and $b_i$ are suppressed and implied when discussing $f_i$. The network prediction is given by 
\begin{align} \label{eqn:F_AE}
F(x; \theta) &= f_{2L} \circ \cdots \circ f_1(x),
\end{align}
where $\theta = \{W_1, \ldots, W_L, b_1, \ldots, b_{2L}\}$ and $x \in \R{n}$. Notice that layers $i$ and $\ind(i)$ both explicitly depend on the parameter $W_i$, for any $i \in \{1, \ldots, L\}$, and their impact on $F$ can be shown by writing $F$ as follows:
\begin{align} \label{eqn:F_AE_expanded}
F(x; \theta) = f_{2L} \circ \cdots \circ f_{\ind(i)} \circ \cdots \circ f_i \circ \cdots \circ f_1(x).
\end{align}
In this section, $\head_i$ and $\tail_i$ are defined analogously to \eqref{eqn:head} and \eqref{eqn:tail} respectively, i.e. 
\begin{align} \label{eqn:head_tail_AE}
\head_i(x) = f_i \circ \cdots \circ f_1(x) \quad \mbox{and} \quad \tail_i(y) = f_{2L} \circ \cdots \circ f_i(y)
\end{align}
for all $x \in \R{n_1}$, $y \in \R{n_i}$, and $i \in \{1, \ldots, 2L\}$. Note again that $\head_0$ and $\tail_{2L+1}$ are identity maps. 

\subsection{Gradient Descent for Standard Loss Function}
For the deep autoencoder, the standard loss function is different. It is of the form 
\begin{align} \label{eqn:std_loss_AE}
J(x; \theta) = \frac{1}{2} \ip{x - F(x;\theta)}{x - F(x;\theta)}.
\end{align}
Notice that the $y$ from \eqref{eqn:standard_loss} has been replaced by $x$ in \eqref{eqn:std_loss_AE}. This is to enforce the output, which is the decoding of the encoded input, to be as similar to the original input as possible. The equation for $\nabla_{\theta_i} J(x; \theta)$ is then updated from the form in \eqref{eqn:dJdT} to 
\begin{align} \label{eqn:dJdT_AE}
\nabla_{\theta_i} J(x; \theta) = \nabla_{\theta_i}^* F(x; \theta) \cdot (F(x; \theta) - x),
\end{align} 
for any parameter $\theta_i$. Note that calculating $\nabla_{W_i}^* F(x;\theta)$ for $i \in \{1, \ldots, L\}$ in this case is more difficult than in \eqref{nabla:Wi:star:F:MLP}, since layers $i$ and $\ind(i)$ both depend on $W_i$. This will be shown towards the end of this section after single-layer derivatives and backpropagation are presented. There is a very strong correspondence between this section and Section \ref{sec:first_order_MLP} because of the similarity in the layerwise-defining function $f_i$, and this will be exploited whenever possible. 

Before proceeding into gradient calculation, however, a very particular instance of the chain rule will be introduced for parameter-dependent maps.

\begin{theorem} \label{theorem:tau_maps}
Let $E, \tilde{E}, H_1, $ and $H_2$ be generic inner product spaces. Consider a linear map $\aelin \in \lin{H_1}{H_2}$, and two parameter-dependent maps $g : E \times H_1 \rightarrow \tilde{E}$ and $h : E \times H_2 \rightarrow \tilde{E}$, such that 
\[
g(x; \theta) = h(x; \aelin(\theta))
\]
for all $x \in E$ and $\theta \in H_1$. Then, the following two results hold for all $U \in H_1$ and $y \in \tilde{E}$
\begin{align*}
\nabla g(x; \theta) \cdot U &= \nabla h(x; \aelin(\theta)) \cdot \aelin(U), \\
\nabla^* g(x; \theta) \cdot y &= \aelin^* \left(\nabla^* h(x; \aelin(\theta))  \cdot y\right).
\end{align*}
\begin{proof}
This is a consequence of the chain rule, the linearity of $\aelin$, and the reversing property of the adjoint. 
\end{proof}
\end{theorem}

Then, single-layer derivatives for a generic function $f$ are presented as corollaries to Theorem \ref{theorem:tau_maps}.

\begin{corollary}
Consider a function $f$ of the form 
\[
f(x; W) = \bignl(\aelin(W) \cdot x + b),
\]
where $x \in \R{n}, b \in \R{m}, W \in \R{n \times m}, \aelin \in \lin{\R{n \times m}}{\R{m \times n}}$, and $\bignl : \R{m} \rightarrow \R{m}$ is an elementwise function. Then, the following hold: for any $U \in \R{n \times m}$, 
\begin{align}
\nabla_W f(x; W) \cdot U &= \D\bignl(z) \cdot \aelin(U) \cdot x, \label{eqn:nabla_W_cor} \\
\nabla_b f(x; W) &= \D\bignl(z), \label{eqn:nabla_b_cor} \\
\D f(x; W) &= \D\bignl(z) \cdot \aelin(W),  \label{eqn:Df_cor}
\end{align}
where $z = \aelin(W) \cdot x + b$. Furthermore, the following hold: for any $y \in \R{m}$,
\begin{align}
\nabla_W^* f(x; W) \cdot y &= \aelin^*\left(\left(\bignl'(z) \odot y \right) x^T \right) \label{eqn:nabla_S_W_cor} \\
\nabla_b^* f(x; W) &= \D\bignl(z) \label{eqn:nabla_S_b_cor} \\
\D^* f(x; W) &= \aelin^*(W) \cdot \D\bignl(z). \label{eqn:Df_S_cor}
\end{align}
\begin{proof}
In Lemmas \ref{lemma:Dfs:MLP} and \ref{lem:D_star_f_MLP}, the derivatives and corresponding adjoints of 
\[
\tilde{f}(x; \tilde{W}) = \bignl(\tilde{W} \cdot x + b)
\]
were calculated, where $\tilde{W} \in \R{m \times n}$. Then, equations \eqref{eqn:nabla_W_cor} and \eqref{eqn:nabla_S_W_cor} are consequences of Lemma \ref{theorem:tau_maps}. 

Equations \eqref{eqn:nabla_b_cor} and \eqref{eqn:Df_cor} also follow from derivatives calculated in Lemmas \ref{lemma:Dfs:MLP} and \ref{lem:D_star_f_MLP}, along with the chain rule. Equations \eqref{eqn:nabla_S_b_cor} and \eqref{eqn:Df_S_cor} follow from the reversing property of the adjoint and the self-adjointness of $\D\bignl(z)$. 
\end{proof}
\end{corollary}

Since the single-layer derivatives can be calculated, it is now shown that backpropagation in a deep autoencoder is of the same form as backpropagation in a MLP.

\begin{theorem}[Backpropagation in Deep AE] \label{thm:backprop_AE} With $f_i$ defined as in \eqref{eqn:AE_f_i} and $\tail_i$ given as in \eqref{eqn:head_tail_AE}, then for any $x_i \in \R{n_i}$ and $i \in \{1, \ldots, 2L\}$,
\[
\D \tail_i(x_i) = \D\tail_{i+1}(x_{i+1}) \cdot \D\bignl_i(z_i) \cdot K_i,
\]
where $z_i = K_i \cdot x_i + b_i$ and $\tail_{L+1}$ is the identity. Furthermore, for any $v \in \R{n_{2L + 1}}$, 
\[
\D^*\tail_i(x_i) \cdot v = K_i^T \cdot \left(\bignl_i'(z_i) \odot \left(\D^*\tail_{i+1}(x_{i+1}) \cdot v\right)\right).
\] 
\begin{proof} 
Since $f_i(x_i) = K_i \cdot x_i + b_i$, where $K_i$ is independent of $x_i$, this result can be proven in the same way as Theorem \ref{thm:MLP_backprop}, replacing $W_i$ with $K_i$.
\end{proof}
\end{theorem}

The derivatives of the entire loss function can now be computed with respect to $W_i$ for any $i \in \{1, \ldots, L\}$, and with respect to $b_i$ for any $i \in \{1, \ldots, 2L\}$.

\begin{theorem} \label{thm:final_first_AE} Let $J$ be defined as in \eqref{eqn:std_loss_AE}, $F$ be defined as in \eqref{eqn:F_AE}, and $\tail_i$ be defined as in \eqref{eqn:head_tail_AE}. Then, for all $i \in \{1, \ldots, L\}$ and $x \in \R{n_1}$, 
\begin{align}
\nabla_{W_i} J(x; \theta)\cdot e &= \left(\bignl_i'(z_i) \odot \left(\D^*\tail_{i+1}(x_{i+1}) \cdot e\right) \right) x_i^T \nonumber \\
& \quad + \aelin_{\ind(i)}^* \left[ \left(\bignl_{\ind(i)}'(z_{\ind(i)}) \odot \left(\D^*\tail_{\ind(i) + 1}(x_{\ind(i)+1}) \cdot e\right) \right) x_{\ind(i)}^T \right], \label{eqn:nabla_W_J_AE}
\end{align}
where $e = F(x; \theta) - x$ and $z_j = K_j \cdot x_j + b_j$ for all $1\leq j\leq 2L$.

Furthermore, for all $i \in \{1, \ldots, 2L\}$, 
\begin{align}
\nabla_{b_i} J(x; \theta) &= \bignl_i'(z_i) \odot \left(\D^*\tail_{i+1}(x_{i+1}) \cdot e\right) \label{eqn:nabla_b_J_AE}.
\end{align}

\begin{proof}
Proving equation \eqref{eqn:nabla_b_J_AE} for any $i \in \{1, \ldots, 2L\}$ is the same as proving \eqref{nabla:b:J:MLP}, and is omitted. 
As for equation \eqref{eqn:nabla_W_J_AE}, recall that only two of the functions comprising $F$ in \eqref{eqn:F_AE_expanded} depend on $W_i$: $f_i$ and $f_{\ind(i)}$. Hence, by the product rule of differentiation, 
\[
\nabla_{W_i}F(x; \theta) 
=  \D\tail_{\ind(i) + 1} (x_{\ind(i) + 1}) \cdot \nabla_{W_i} f_{\ind(i)} (x_{\ind(i)}) + \D\tail_{i+1}(x_{i+1}) \cdot \nabla_{W_i} f_i(x_i).
\]
Taking the adjoint of this implies
\begin{align}\label{nabla:W:F:star:AE}
\nabla^*_{W_i} F(x;\theta) \cdot e = \nabla_{W_i}^*f_{\ind(i)}(x_{\ind(i)}) \cdot \D^*\tail_{\ind(i)+1}(x_{\ind(i)+1}) \cdot e+ \nabla_{W_i}^*f_i(x_i) \cdot \D^*\tail_{i+1}(x_{i+1}) \cdot e.
\end{align}
Equation \eqref{nabla:star:Wi:f:MLP} gives
\begin{equation}\label{nab:star:f:AE}
\nabla^*_{W_i}f_i(x_i) \cdot u = \left(\bignl_i'(z_i) \odot u\right) x_i^T
\end{equation}
for any $u \in \R{n_{i+1}}$  and any $i \in \{1, \ldots, L\}$.  Since $i \in \{1, \ldots, L\}$ implies $\ind(i) \in \{L+1, \ldots, 2L\}$,  equation \eqref{eqn:nabla_S_W_cor}  implies
\begin{equation}\label{nab:star:xi:f:AE}
\nabla^*_{W_i} f_{\ind(i)}(x_{\ind(i)})\cdot v = \aelin_{\ind(i)}^*\left(\left(\bignl_{\ind(i)}'(z_{\ind(i)}) \odot v\right) x_{\ind(i)}^T\right)
\end{equation}
for any $v \in \R{\ind(i)+1}$ and any $i \in \{1, \ldots, L\}$, where $z_{\ind(i)} = \aelin_{\ind(i)} (W_i) \cdot x_{\ind(i)} + b_{\ind(i)}$. Hence,  \eqref{eqn:nabla_W_J_AE} follows from \eqref{eqn:dJdT_AE} and \eqref{nabla:W:F:star:AE} -- \eqref{nab:star:xi:f:AE}.
\end{proof}
\end{theorem}

One iteration of a gradient descent algorithm to minimize $J$ with respect to the parameters is given in Algorithm \ref{alg:AE_grad_desc_1}. As before, the output of this algorithm is a new parameter set $\theta = \{W_1, \ldots W_L, b_1, \ldots, b_{2L}\}$ that has taken one step in the direction of the negative gradient of $J$ with respect to each parameter. 

\begin{algorithm} 
\caption{One iteration of gradient descent in an autoencoder}
\label{alg:AE_grad_desc_1}
\begin{algorithmic}
\Function{Descent Iteration}{$x, W_1, \ldots, W_L, b_1, \ldots, b_{2L}, \eta$}
\State $x_1 \gets x$
\For {$i \in \{1, \ldots, 2L\}$} \Comment $x_{2L+1} = F(x;\theta)$
\If {$i <= L$}
\State $K_i \gets W_i$
\Else
\State $K_i \gets \aelin_i(W_{\ind(i)})$
\EndIf
\State $z_i \gets K_i \cdot x_i + b_i$
\State $x_{i+1} \gets \bignl_i(z_i)$
\EndFor
\For {$i \in \{2L, \ldots, 1\}$}
\If {$i = 2L$} \Comment $\tail_{2L+1} = \mbox{identity}$
\State $e_x \gets x_{2L+1} - x$ \Comment $e_x = \D^*\tail_{i+1}(x_{i+1})\cdot \left(x_{2L+1} - x\right)$
\Else
\State $e_x \gets K_{i+1}^T \cdot \left(\bignl_i'(z_{i+1}) \odot e_x\right)$ \Comment Thm. \ref{thm:backprop_AE}
\EndIf
\State $\nabla_{b_i} J(x; \theta) \gets \bignl_i'(z_i) \odot e_x$ \Comment \eqref{eqn:nabla_b_J_AE}
\State $b_i \gets b_i - \eta \nabla_{b_i} J(x;\theta)$
\If {$i > L$}
\State $\nabla_{W_{\ind(i)}} J(x; \theta) \gets \aelin_i^*\left(\left(\bignl_i'(z_i) \odot e_x\right) x_i^T\right)$ \Comment Second term in \eqref{eqn:nabla_W_J_AE}
\Else
\State $\nabla_{W_i}J(x; \theta) \gets \nabla_{W_i}J(x; \theta) + \left(\bignl_i'(z_i) \odot e_x\right) x_i^T$ \Comment Add first term in \eqref{eqn:nabla_W_J_AE}
\State $W_i \gets W_i - \eta \nabla_{W_i} J(x;\theta)$
\EndIf
\EndFor
\EndFunction
\end{algorithmic}
\end{algorithm}

\subsection{Gradient Descent for Higher-Order Loss Function}
Now, as in previous sections, a loss function $\mathcal{J}_R = J + \mu R$ is considered, with $R(x; \theta)$ defined as in \eqref{eqn:R} or \eqref{eqn:R_mult_terms}. To perform gradient descent to minimize $\mathcal{J}_R$, it is only necessary to determine the gradient of $R$ with respect to the parameters, since the gradient of $J$ can already be calculated. Again, forward and backward propagation through the tangent network must be computed in the spirit of \cite{simard1992tangent}, as well as $(\nabla \D F(x; \theta) \righthook v)^*$.

\begin{lemma} \label{lem:forward_prop_AE}
For $f_i$ defined as in \eqref{eqn:AE_f_i}, $\head_i$ defined as in \eqref{eqn:head_tail_AE}, and any $x, v \in \R{n}$, 
\[
\D\head_i(x) \cdot v = \bignl'_i(z_i) \odot \left(K_i \cdot \D\head_{i-1}(x) \cdot v\right),
\]
where $x_i = \head_{i-1}(x)$ and $z_i = K_i \cdot x_i + b_i$, for all $i \in \{1, \ldots, 2L\}$. 
\begin{proof}
This result is proven similarly to Theorem \ref{theorem:FP:alpha:high:MLP} since $f_i(x_i) = \bignl_i(K_i \cdot x_i + b_i)$.
\end{proof}
\end{lemma}

Now, tangent backpropagation must be computed. 
\begin{theorem} [Tangent Backpropagation in Deep AE] \label{thm:tgt_backprop_AE} Let $\head_i$ and $\tail_i$ be defined as in \eqref{eqn:head_tail_AE}. Let $f_i$ be defined as in \eqref{eqn:AE_f_i}. Then, for any $i \in \{1, \ldots, 2L\}$, $x, v_1 \in \R{n_1},$ and $v_2 \in \R{n_{2L+1}}$, 
\begin{align*}
\left( (\D \head_{i-1}(x) \cdot v_1) \lefthook \D^2\tail_i(x_i)\right)^*\! \cdot\! v_2 &= K_i^T \cdot \left\{\bignl_i'(z_i) \odot \left[ \left((\D\head_i(x) \cdot v_1) \lefthook \D^2\tail_{i+1}(x_{i+1})\right)^* \cdot v_2\right] \right\} \\
&+ K_i^T\! \cdot\! \left\{ \bignl''_i(z_i) \odot (K_i \!\cdot\! \D\head_{i-1}(x) \cdot v_1) \odot (\D^*\tail_{i+1}(x_{i+1}) \!\cdot\! v_2) \right\},
\end{align*}
where $x_i = \head_{i-1}(x)$ and $z_i = K_i \cdot x_i + b_i$. Also, 
\[
\left( (\D\head_{2L}(x) \cdot v_1) \lefthook \D^2 \tail_{2L+1}(x_{2L+1})\right)^* \cdot v_2 = 0.
\]
\begin{proof}
Since $f_i(x_i) = \bignl_i(K_i \cdot x_i + b_i)$ and $K_i$ is independent of $x_i$, this result can be proven in the same way as Theorem \ref{theorem:tangent:BP:MLP}. 
\end{proof}
\end{theorem}

Since the tangents can be backpropagated, the final step in calculating the gradients of $R$ is to calculate $(\nabla_{\theta_i} \D F(x;\theta) \righthook v)^*$, where $\theta_i$ is a generic parameter. 

\begin{theorem} Let $\head_i$ and $\tail_i$ be defined in \eqref{eqn:head_tail_AE}, and $F$ be defined as in \eqref{eqn:F_AE}. Then, for any $e \in \R{n_{2L+1}}$, $x \in \R{n_1}$, and $i \in \{1, \ldots, L\}$,
\begin{align}
&\left(\nabla_{W_i} \D F(x;\theta) \righthook v\right)^* \cdot e \nonumber \\ \quad &= \left\{\bignl_i'(z_i) \odot \left[\left((\D \head_i(x)\cdot v) \lefthook \D^2\tail_{i+1}(x_{i+1})\right)^* \cdot e\right]\right\} x_i^T \nonumber \\
 &\qquad + \left(\bignl''_i(z_i) \odot (K_i \cdot \D\head_{i-1}(x) \cdot v) \odot (\D^*\tail_{i+1}(x_{i+1}) \cdot e) \right)x_i^T \nonumber \\
 &\qquad + \left(\bignl_i'(z_i) \odot (\D^*\tail_{i+1}(x_{i+1}) \cdot e)\right) (\D\head_{i-1}(x)\cdot v)^T \label{eqn:nabla_W_D_F_AE} \\
\quad &+ \aelin_{\ind(i)}^* \left(\left\{\bignl_{\ind(i)}'(z_{\ind(i)}) \odot \left[\left((\D \head_{\ind(i)}(x)\cdot v) \lefthook \D^2\tail_{\ind(i)+1}(x_{\ind(i)+1})\right)^* \cdot e\right]\right\} x_{\ind(i)}^T\right) \nonumber \\
&\qquad + \aelin_{\ind(i)}^* \left[\left(\bignl''_{\ind(i)}(z_{\ind(i)}) \odot (K_{\ind(i)} \cdot \D\head_{\ind(i)-1}(x) \cdot v) \odot (\D^*\tail_{\ind(i)+1}(x_{\ind(i)+1}) \cdot e) \right)x_{\ind(i)}^T\right] \nonumber \\
&\qquad + \aelin_{\ind(i)}^* \left[\left(\bignl_{\ind(i)}'(z_{\ind(i)}) \odot (\D^*\tail_{\ind(i)+1}(x_{\ind(i)+1}) \cdot e)\right) (\D\head_{\ind(i)-1}(x)\cdot v)^T\right], \nonumber
\end{align}
where $x_i = \head_{i-1}(x)$ and $z_i = K_i \cdot x_i + b_i$. Furthermore, for any $i \in \{1, \ldots, 2L\}$, 
\begin{align}
\left(\nabla_{b_i} \D F(x;\theta) \righthook v\right)^* \cdot e &= \bignl_i'(z_i) \odot \left[\left((\D \head_i(x)\cdot v) \lefthook \D^2\tail_{i+1}(x_{i+1})\right)^* \cdot e \right] \nonumber \\
&\qquad + \bignl_i''(z_i) \odot (K_i \cdot \D\head_{i-1}(x)\cdot v) \odot \left(\D^*\tail_{i+1}(x_{i+1}) \cdot e \right). \label{eqn:nabla_b_D_F_AE}
\end{align}

\begin{proof}
Equation \eqref{eqn:nabla_b_D_F_AE} is proven similarly to \eqref{eqn:nabla_b_D_F_MLP} and is omitted. Equation \eqref{eqn:nabla_W_D_F_AE} is now derived.

Consider the case when $i \in \{1, \ldots, L\}$. Recall from the proof of Theorem \ref{thm:final_first_AE} that 
\[
\nabla_{W_i} F(x; \theta) = \D\tail_{i+1}(x_{i+1}) \cdot \nabla_{W_i} f_i(x_i) + \D\tail_{\ind(i) + 1} (x_{\ind(i) + 1}) \cdot \nabla_{W_i} f_{\ind(i)}(x_{\ind(i)}).
\]
Then, as in the proof of Theorem \ref{thm:DF_righthook_star},
\begin{align} \label{eqn:final_second_AE_tmep}
\left(\nabla_{W_i} \D F(x; \theta) \righthook v\right) &= \left( (\D\head_i(x) \cdot v) \lefthook \D^2\tail_{i+1}(x_{i+1}) \right) \cdot \nabla_{W_i} f_i(x_i) \\
&\qquad + \D\tail_{i+1}(x_{i+1}) \cdot \left( (\D\head_{i-1}(x) \cdot v) \lefthook \D\nabla_{W_i} f_i(x_i) \right) \nonumber \\
&+ \left((\D\head_{\ind(i)} \cdot v) \lefthook \D^2\tail_{\ind(i)+1}(x_{\ind(i)+1}) \right) \cdot \nabla_{W_i} f_{\ind(i)}(x_{\ind(i)}) \nonumber \\
&\qquad + \D\tail_{\ind(i)+1}(x_{\ind(i)+1}) \cdot \left((\D\head_{\ind(i)-1}(x) \cdot v) \lefthook \D\nabla_{W_i} f_{\ind(i)}(x_{\ind(i)})\right), \nonumber
\end{align}
where the third and fourth terms come from the second term in $\nabla_{W_i} F(x; \theta)$. Then, taking the adjoint of the first two terms of \eqref{eqn:final_second_AE_tmep} works as in \eqref{eqn:nabla_W_D_F_MLP}, replacing $W_i$ with $K_i$. Taking the adjoint of the final two terms of \eqref{eqn:final_second_AE_tmep} can be done using Theorem \ref{theorem:tau_maps} and \eqref{eqn:nabla_W_D_F_MLP}, which completes the proof.
\end{proof}
\end{theorem}

\begin{corollary}
Let $\head_i$ and $\tail_i$ be defined in \eqref{eqn:head_tail_AE}, and $F$ be defined as in \eqref{eqn:F_AE}. Then, for any $e \in \R{n_{2L+1}}$, $x \in \R{n_1}$, and $i \in \{1, \ldots, L\}$,
\begin{align}
&\left(\nabla_{W_i} \D F(x;\theta) \righthook v\right)^* \cdot e \nonumber \\  &\qquad= \left(\left(\nabla_{b_i} \D F(x; \theta) \righthook v\right)^*\cdot e\right) x_i^T \nonumber \\ &\quad \qquad+ \left(\bignl_i'(z_i) \odot (\D^*\tail_{i+1}(x_{i+1}) \cdot e)\right) (\D\head_{i-1}(x)\cdot v)^T  \label{eqn:nabla_W_D_F_cor} \\
&\quad \qquad + \aelin_{\ind(i)}^* \left[\left(\left(\nabla_{b_{\ind(i)}} \D F(x; \theta) \righthook v\right)^*\cdot e\right) x_{\ind(i)}^T \right] \nonumber \\
&\quad \qquad + \aelin_{\ind(i)}^* \left[\left(\bignl_{\ind(i)}'(z_{\ind(i)}) \odot (\D^*\tail_{\ind(i)+1}(x_{\ind(i)+1}) \cdot e)\right) (\D\head_{\ind(i)-1}(x)\cdot v)^T\right] \nonumber,
\end{align}
where $x_i = \head_{i-1}(x)$ and $z_i = K_i \cdot x_i + b_i$.
\begin{proof}
This result can easily be obtained by substituting \eqref{eqn:nabla_b_D_F_AE} into \eqref{eqn:nabla_W_D_F_AE}.
\end{proof}
\end{corollary}

Recall the following for $v_x, \beta_x \in \R{n_{2L+1}}$:
\[
\nabla_{\theta_i} R(x;\theta) = \left(\nabla_{\theta_i} \D F(x;\theta) \righthook v_x\right)^* \cdot (\D F(x;\theta) \cdot v_x - \beta_x), 
\]
for a generic parameter $\theta_i$. Now, gradient descent can be performed to minimize $\mathcal{J}_R = J + \mu R$ since the gradient of $R$ is known. One iteration of this is given in Algorithm \ref{alg:AE_grad_desc_2}. 

\begin{algorithm}
\caption{One iteration of gradient descent for higher-order loss in an autoencoder}
 \label{alg:AE_grad_desc_2}
\begin{algorithmic}
\Function{Descent Iteration}{$x, v_x, \beta_x, W_1, \ldots, W_L, b_1, \ldots, b_{2L}, \eta, \mu$}
\State $x_1 \gets x$ 
\State $v_1 \gets v_x$ \Comment $v_i = \D\head_{i-1}(x) \cdot v_x$ and $\D\head_0(x) = $ identity
\For {$i \in \{1, \ldots, 2L\}$} \Comment $x_{2L+1} = F(x; \theta)$ and $v_{2L+1} = \D F(x; \theta) \cdot v_x$
\If {$i <= L$}
\State $K_i \gets W_i$
\Else
\State $K_i \gets \aelin_i(W_{\ind(i)})$
\EndIf
\State $z_i \gets K_i \cdot x_i + b_i$
\State $x_{i+1} \gets \bignl_i(z_i)$
\State $v_{i+1} \gets \bignl_i'(z_i) \odot (K_i \cdot v_i)$ \Comment Lemma \ref{lem:forward_prop_AE}
\EndFor
\For {$i \in \{2L, \ldots, 1\}$}
\If {$i = 2L$} \Comment $\tail_{2L+1} = \mbox{identity}$
\State $e_x \gets x_{2L+1} - x$ \Comment $e_x = \D^*\tail_{i+1}(x_{i+1})\cdot \left(x_{2L+1} - x\right)$
\State $e_t \gets 0$ \Comment $e_t = \left(v_{i+1} \lefthook \D^2 \tail_{i+1}(x_{i+1})\right)^* \cdot \left(v_{2L+1} - \beta_x\right)$
\State $e_v \gets v_{2L+1} - \beta_x$ \Comment $e_v = \D^*\tail_{i+1}(x_{i+1}) \cdot (v_{2L+1} - \beta_x)$
\Else
\State $e_x \gets K_{i+1}^T \cdot \left(\bignl_{i+1}'(z_{i+1}) \odot e_x\right)$ \Comment Thm. \ref{thm:backprop_AE}
\State $e_t \gets K_{i+1}^T \cdot \left(\bignl_{i+1}'(z_{i+1}) \odot e_t + \bignl_{i+1}''(z_{i+1}) \odot (K_{i+1} \cdot v_{i+1}) \odot e_v\right)$ \Comment Thm. \ref{thm:tgt_backprop_AE}; old $e_v$ 
\State $e_v \gets K_{i+1}^T \cdot \left(\bignl_{i+1}'(z_{i+1}) \odot e_v\right)$ \Comment Thm. \ref{thm:backprop_AE}
\EndIf
\State $\nabla_{b_i} J(x; \theta) \gets \bignl_i'(z_i) \odot e_x$ \Comment \eqref{eqn:nabla_b_J_AE}
\State $\nabla_{b_i} R(x; \theta) \gets \bignl_i'(z_i) \odot e_t + \bignl_i''(z_i) \odot (K_i \cdot v_i) \odot e_v$ \Comment \eqref{eqn:nabla_b_D_F_AE}
\State $b_i \gets b_i - \eta \left(\nabla_{b_i} J(x;\theta) + \mu \nabla_{b_i} R(x; \theta) \right)$ 
\If {$i > L$}
\State $\nabla_{W_{\ind(i)}} J(x; \theta) \gets \aelin_i^*\left(\left(\bignl_i'(z_i) \odot e_x\right) x_i^T\right)$ \Comment Second term in \eqref{eqn:nabla_W_J_AE}
\State $\nabla_{W_{\ind(i)}} R(x; \theta) \gets \aelin_i^* \left( \left(\nabla_{b_i} R(x; \theta)\right) x_i^T + \left(\bignl_i'(z_i) \odot e_v\right) v_i^T \right)$ \Comment Terms 3 \& 4 in \eqref{eqn:nabla_W_D_F_cor}
\Else
\State $\nabla_{W_i}J(x; \theta) \gets \nabla_{W_i}J(x; \theta) + \left(\bignl_i'(z_i) \odot e_x\right) x_i^T$ \Comment Add first term in \eqref{eqn:nabla_W_J_AE}
\State $\nabla_{W_i} R(x; \theta) \gets \nabla_{W_i} R(x; \theta) +  \left(\nabla_{b_i} R(x; \theta)\right) x_i^T + \left(\bignl_i'(z_i) \odot e_v\right) v_i^T$ 
\State \Comment Terms 1 \& 2 in  \eqref{eqn:nabla_W_D_F_cor}, add to previously computed result. 
\State $W_i \gets W_i - \eta \left(\nabla_{W_i} J(x;\theta) + \mu \nabla_{W_i} R(x; \theta)\right) $
\EndIf
\EndFor
\EndFunction
\end{algorithmic}
\end{algorithm}

\section{Conclusion and Future Work}
In this work, a concise and complete mathematical framework for DNNs was formulated. Generic multivariate functions defined the operation of the network at each layer, and their composition defined the overall mechanics of the network. A coordinate-free gradient descent algorithm, which relied heavily on derivatives of vector-valued functions, was presented and applied to two specific examples. It was shown how to calculate gradients of network loss functions over the inner product space in which the parameters reside, as opposed to individually with respect to each component. A simple loss function and a higher-order loss function were considered, and it was also shown how to extend this framework to other types of loss functions. The approach considered in this paper was generic and flexible and can be extended to other types of networks besides the ones considered here. 

The most immediate direction of future work would be to represent the parameters of a DNN in some sort of lower-dimensional subspace to promote sparsity in the network. Finding meaningful basis representations of parameters could help limit the amount of overfitting, while still maintaining the predictive power of the model. Also, more sophisticated optimization methods become tractable once the number of dimensions is sufficiently reduced, and it would be interesting to apply these to neural networks. Another direction for future work is to exploit the discrete-time dynamical system structure presented for the layerwise network, and to consider how to use control and dynamical systems theory to improve network training or output.

\bibliographystyle{plain}
\bibliography{refs}

\end{document}